\documentclass{article} 
\usepackage{iclr2020_conference,times}

\usepackage{amsmath,amsfonts,bm}

\def\eqref#1{equation~\ref{#1}}

\def\1{\bm{1}}

\DeclareMathAlphabet{\mathsfit}{\encodingdefault}{\sfdefault}{m}{sl}
\SetMathAlphabet{\mathsfit}{bold}{\encodingdefault}{\sfdefault}{bx}{n}

\newcommand{\myparagraph}[1]{\vspace{-3pt}\paragraph{#1}}

\newcommand{\motherNet}{once-for-all network}
\newcommand{\motherNetCap}{Once-for-all Network}
\newcommand{\childNet}{sub-network}
\newcommand{\childNetCap}{Sub-network}

\definecolor{MyDarkBlue}{rgb}{0,0.08,1}
\definecolor{MyDarkGreen}{rgb}{0.02,0.6,0.02}
\definecolor{MyDarkRed}{rgb}{0.8,0.02,0.02}
\definecolor{MyDarkOrange}{rgb}{0.40,0.2,0.02}
\definecolor{MyPurple}{RGB}{111,0,255}
\definecolor{MyRed}{rgb}{1.0,0.0,0.0}
\definecolor{MyGold}{rgb}{0.75,0.6,0.12}
\definecolor{MyDarkgray}{rgb}{0.66, 0.66, 0.66}

\usepackage{color,xcolor}
\usepackage{epsfig}
\usepackage{graphicx}

\usepackage{adjustbox}
\usepackage{array}
\usepackage{booktabs}
\usepackage{colortbl}
\usepackage{float,wrapfig}
\usepackage{hhline}
\usepackage{multirow}
\usepackage{subcaption} 

\usepackage{amsmath,amsfonts,amsthm,amssymb}
\usepackage{bm}
\usepackage{nicefrac}
\usepackage{microtype}

\usepackage{changepage}
\usepackage{extramarks}
\usepackage{fancyhdr}
\usepackage{lastpage}
\usepackage{setspace}
\usepackage{soul}
\usepackage{xspace}

\usepackage[pagebackref=true,breaklinks=true,colorlinks,citecolor=gray]{hyperref}
\usepackage{url}
\usepackage{natbib}

\usepackage{titlesec}
\usepackage{listings}
\usepackage{enumitem}

\usepackage[utf8]{inputenc}

\title{Once-for-All: Train One Network and Specialize it for Efficient Deployment}

\author{
    Han Cai$^1$, Chuang Gan$^2$, Tianzhe Wang$^1$, Zhekai Zhang$^1$, Song Han$^1$\\
    $^1$Massachusetts Institute of Technology, \quad $^2$MIT-IBM Watson AI Lab\\
    {\small\texttt{\{hancai, chuangg, songhan\}@mit.edu}}
}

\iclrfinalcopy 
\begin{document}

\maketitle
\begin{abstract}
We address the challenging problem of efficient inference across many devices and resource constraints, especially on edge devices.  Conventional approaches either manually design or use neural architecture search (NAS) to find a specialized neural network and train it from scratch for \textit{each} case, which is computationally prohibitive (causing $CO_2$ emission as much as 5 cars' lifetime \cite{strubell2019energy}) thus unscalable. 
In this work, we propose to train a once-for-all (OFA) network that supports diverse architectural settings by decoupling training and search, to reduce the cost. 
We can quickly get a specialized sub-network by selecting from the OFA network without additional training. To efficiently train OFA networks, we also propose a novel progressive shrinking algorithm, a generalized pruning method that reduces the model size across many more dimensions than pruning (depth, width, kernel size, and resolution). It can obtain a surprisingly large number of sub-networks ($> 10^{19}$) that can fit different hardware platforms and latency constraints while maintaining the same level of accuracy as training independently. 
On diverse edge devices, OFA consistently outperforms state-of-the-art (SOTA) NAS methods (up to 4.0\% ImageNet top1 accuracy improvement over MobileNetV3, or same accuracy but 1.5$\times$ faster than MobileNetV3, 2.6$\times$ faster than EfficientNet w.r.t measured latency) while reducing many orders of magnitude GPU hours and $CO_2$ emission. In particular, OFA achieves a new SOTA 80.0\% ImageNet top-1 accuracy under the mobile setting ($<$600M MACs). OFA is the winning solution for the 3rd Low Power Computer Vision Challenge (LPCVC), DSP classification track and the 4th LPCVC, both classification track and detection track. Code and  50 pre-trained models (for many devices \& many latency constraints) are released at \url{https://github.com/mit-han-lab/once-for-all}. 
\end{abstract}

\section{Introduction}

Deep Neural Networks (DNNs) deliver state-of-the-art accuracy in many machine learning applications.
However, the explosive growth in model size and computation cost gives rise to new challenges on how to efficiently deploy these deep learning models on \emph{diverse} hardware platforms, since they have to meet \emph{different} hardware efficiency constraints (e.g., latency, energy). 
For instance, one mobile application on App Store has to support a diverse range of hardware devices, from a high-end Samsung Note10 with a dedicated neural network accelerator to a 5-year-old Samsung S6 with a much slower processor. With different hardware resources (e.g., on-chip memory size, \#arithmetic units), the optimal neural network architecture varies significantly. Even running on the same hardware, under different battery conditions or workloads, the best model architecture also differs a lot.

\begin{figure}[t]
    \vspace{-5pt}
    \centering
    \includegraphics[width=1\linewidth]{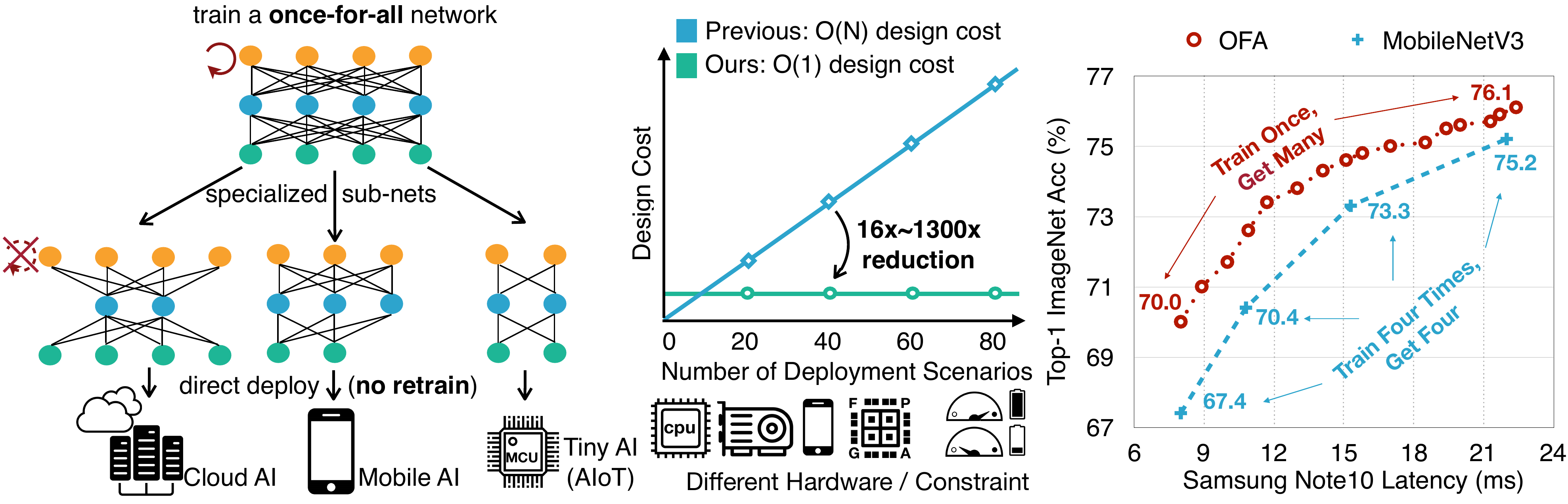}
    \caption{Left: a single \motherNet~is trained to support versatile architectural configurations including depth, width, kernel size, and resolution. Given a deployment scenario, a specialized sub-network is directly selected from the \motherNet~without training. Middle: this approach reduces the cost of specialized deep learning deployment from O(N) to O(1). Right: once-for-all network followed by model selection can derive many accuracy-latency trade-offs by training only once, compared to conventional methods that require repeated training. 
    }\label{fig:large_scale_deep_learning_deployment}
\end{figure}

Given different hardware platforms and efficiency constraints (defined as deployment scenarios), researchers either design compact models specialized for mobile~\citep{howard2017mobilenets,sandler2018mobilenetv2,zhang2018shufflenet} or accelerate the existing models by compression \citep{han2016deep, he2018amc} for efficient deployment. However, designing specialized DNNs for every scenario is engineer-expensive and computationally expensive, either with human-based methods or NAS. Since such methods need to \emph{repeat} the network design process and \emph{retrain} the designed network from scratch for \emph{each} case. Their total cost grows linearly as the number of deployment scenarios increases, which will result in excessive energy consumption and $CO_2$ emission \citep{strubell2019energy}. It makes them unable to handle the vast amount of hardware devices (23.14 billion IoT devices till 2018\footnote{https://www.statista.com/statistics/471264/iot-number-of-connected-devices-worldwide/}) and highly dynamic deployment environments (different battery conditions, different latency requirements, etc.). 

This paper introduces a new solution to tackle this challenge -- designing a \textit{\motherNet}~that can be directly deployed under diverse architectural configurations, amortizing the training cost. The inference is performed by selecting only part of the \motherNet. It flexibly supports different depths, widths, kernel sizes, and resolutions without retraining.
A simple example of \emph{Once-for-All} (OFA) is illustrated in Figure~\ref{fig:large_scale_deep_learning_deployment} (left). Specifically, we decouple the model training stage and the neural architecture search stage. In the model training stage, we focus on improving the accuracy of all sub-networks that are derived by selecting different parts of the \motherNet. 
In the model specialization stage, we sample a subset of sub-networks to train an accuracy predictor and latency predictors. Given the target hardware and constraint, a predictor-guided architecture search \citep{liu2018progressive} is conducted to get a specialized sub-network, and the cost is negligible.  As such, we reduce the total cost of specialized neural network design from O(N) to O(1) (Figure~\ref{fig:large_scale_deep_learning_deployment} middle). 

However, training the \motherNet~is a non-trivial task, since it requires joint optimization of the weights to maintain the accuracy of a large number of sub-networks (more than 10$^{19}$ in our experiments). It is computationally prohibitive to enumerate all sub-networks to get the exact gradient in each update step, while randomly sampling a few sub-networks in each step will lead to significant accuracy drops. The challenge is that different sub-networks are interfering with each other, making the training process of the whole \motherNet~inefficient. 
To address this challenge, we propose a \textit{progressive shrinking} algorithm for training the \motherNet. Instead of directly optimizing the \motherNet~from scratch, we propose to first train the largest neural network with \emph{maximum} depth, width, and kernel size, then progressively fine-tune the \motherNet~to support \textit{smaller} sub-networks that share weights with the larger ones. As such, it provides better initialization by selecting the most important weights of larger sub-networks, and the opportunity to distill smaller sub-networks, which greatly improves the training efficiency. From this perspective, progressive shrinking can be viewed as a generalized network pruning method that shrinks multiple dimensions (depth, width, kernel size, and resolution) of the full network rather than only the width dimension. Besides, it targets on maintaining the accuracy of all sub-networks rather than a single pruned network. 

\begin{figure}[t]
    \vspace{-20pt}
    \centering
    \includegraphics[width=1\linewidth]{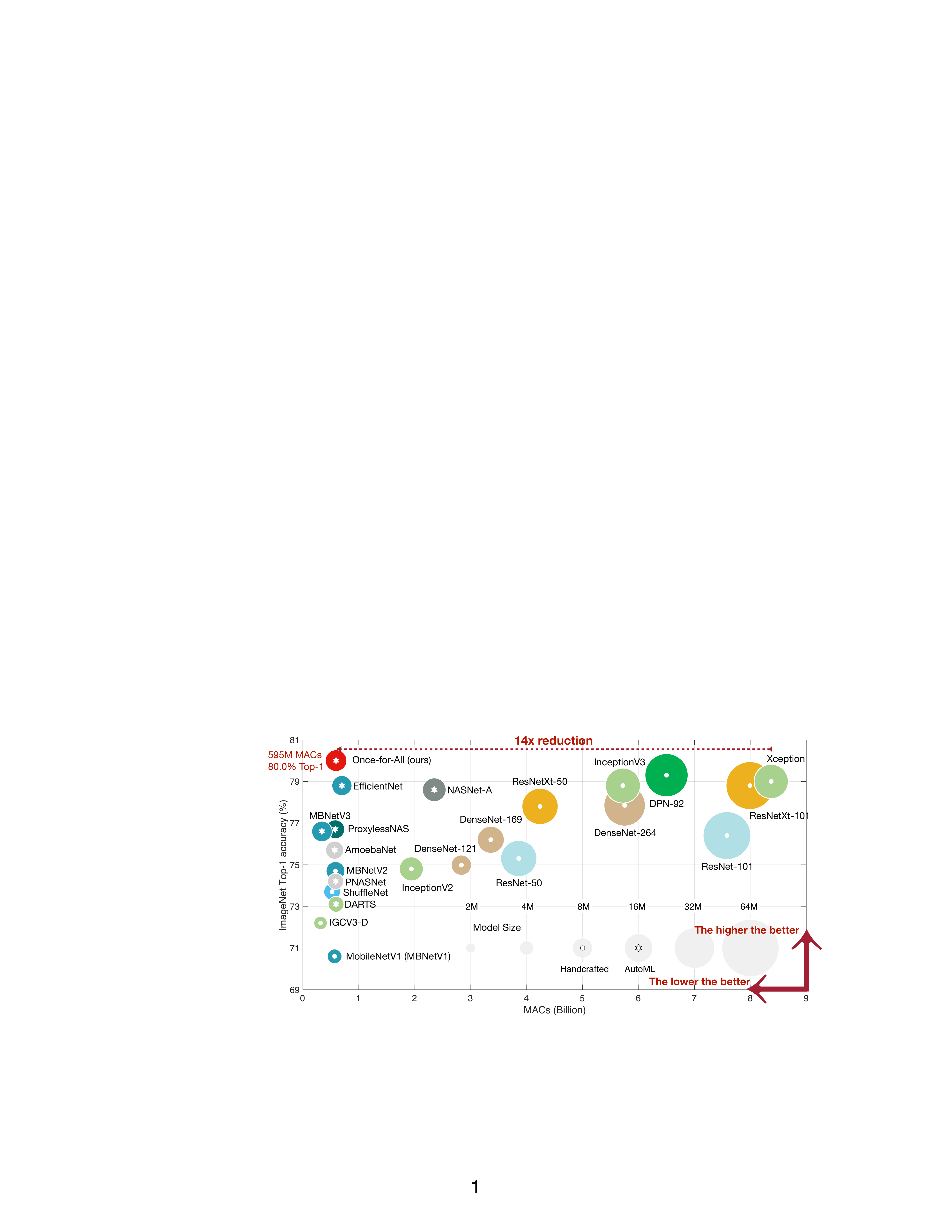}
    \caption{Comparison between OFA and state-of-the-art CNN models on ImageNet. OFA provides 80.0\% ImageNet top1 accuracy under the mobile setting ($<$ 600M MACs).}\label{fig:cnn_imagenet}
\end{figure}

We extensively evaluated the effectiveness of OFA on ImageNet with many hardware platforms (CPU, GPU, mCPU, mGPU, FPGA accelerator) and efficiency constraints. Under all deployment scenarios, OFA consistently improves the ImageNet accuracy by a significant margin compared to SOTA hardware-aware NAS methods while saving the GPU hours, dollars, and $CO_2$ emission by orders of magnitude. On the ImageNet mobile setting (less than 600M MACs), OFA achieves a new SOTA 80.0\% top1 accuracy with 595M MACs (Figure~\ref{fig:cnn_imagenet}). To the best of our knowledge, this is the first time that the SOTA ImageNet top1 accuracy reaches 80\% under the mobile setting.

\section{Related Work}
\myparagraph{Efficient Deep Learning.} 
Many efficient neural network architectures are proposed to improve the hardware efficiency, such as SqueezeNet \citep{iandola2016squeezenet}, MobileNets \citep{howard2017mobilenets,sandler2018mobilenetv2}, ShuffleNets \citep{ma2018shufflenet,zhang2018shufflenet}, etc. 
Orthogonal to architecting efficient neural networks, model compression \citep{han2016deep} is another very effective technique for efficient deep learning,
including network pruning that removes redundant units \citep{han2015learning} or redundant channels \citep{he2018amc,liu2017learning}, and quantization that reduces the bit width for the weights and activations \citep{han2016deep,courbariaux2015binaryconnect,zhu2016trained}. 

\myparagraph{Neural Architecture Search.} 
Neural architecture search (NAS) focuses on automating the architecture design process \citep{zoph2017neural,zoph2018learning,real2018regularized,cai2018efficient,liu2019darts}.
Early NAS methods \citep{zoph2018learning,real2018regularized,cai2018path} search for high-accuracy architectures without taking hardware efficiency into consideration. Therefore, the produced architectures (e.g., NASNet, AmoebaNet) are not efficient for inference. Recent hardware-aware NAS methods \citep{cai2019proxylessnas, tan2018mnasnet, wu2018fbnet} directly incorporate the hardware feedback into architecture search. Hardware-DNN co-design techniques \citep{jiang2019accuracy,jiang2019hardware,hao2019fpga} jointly optimize neural network architectures and hardware architectures. 
As a result, they can improve inference efficiency. 
However, given new inference hardware platforms, these methods need to repeat the architecture search process and retrain the model, leading to prohibitive GPU hours, dollars, and $CO_2$ emission. They are not scalable to a large number of deployment scenarios. The individually trained models do not share any weight, leading to large total model size and high downloading bandwidth.

\myparagraph{Dynamic Neural Networks.} 
To improve the efficiency of a given neural network, some work explored skipping part of the model based on the input image. For example, \cite{wu2018blockdrop,liu2018dynamic,wang2018skipnet} learn a controller or gating modules to adaptively drop layers; \cite{huang2017multi} introduce early-exit branches in the computation graph;
\cite{lin2017runtime} adaptively prune channels based on the input feature map; \cite{kuen2018stochastic} introduce stochastic downsampling point to reduce the feature map size adaptively. Recently, Slimmable Nets \citep{yu2018slimmable,yu2019universally} propose to train a model to support multiple width multipliers (e.g., 4 different global width multipliers), building upon existing human-designed neural networks (e.g., MobileNetV2 0.35, 0.5, 0.75, 1.0). Such methods can adaptively fit different efficiency constraints at runtime, however, still inherit a pre-designed neural network (e.g., MobileNet-v2), which limits the degree of flexibility (e.g., only width multiplier can adapt) and the ability in handling new deployment scenarios where the pre-designed neural network is not optimal. In this work, in contrast, we enable a much more diverse architecture space (depth, width, kernel size, and resolution) and a significantly larger number of architectural settings ($10^{19}$ v.s. 4 \citep{yu2018slimmable}). Thanks to the diversity and the large design space, we can derive new specialized neural networks for many different deployment scenarios rather than working on top of an existing neural network that limits the optimization headroom. However, it is more challenging to train the network to achieve this flexibility, which motivates us to design the progressive shrinking algorithm to tackle this challenge. 

\section{Method}

\subsection{Problem Formalization}
Assuming the weights of the \motherNet~as $W_o$ and the architectural configurations as $\{arch_i\}$, we then can formalize the problem as 
\begin{equation}\small
\label{eq:objective}
    \min_{W_o} \sum_{arch_i} \mathcal{L}_{val} \big( C(W_o, arch_i) \big),
\end{equation}
where $C(W_o, arch_i)$ denotes a selection scheme that selects part of the model from the \motherNet~$W_o$ to form a sub-network with architectural configuration $arch_i$. The overall training objective is to optimize $W_o$ to make each supported sub-network maintain the \textit{same} level of accuracy as \textit{independently} training a network with the same architectural configuration.

\subsection{Architecture Space}
Our \motherNet~provides one model but supports many sub-networks of different sizes, covering four important dimensions of the convolutional neural networks (CNNs) architectures, i.e., depth, width, kernel size, and resolution.  
Following the common practice of many CNN models \citep{he2016deep,sandler2018mobilenetv2,huang2017densely}, we divide a CNN model into a sequence of units with gradually reduced feature map size and increased channel numbers. Each unit consists of a sequence of layers where only the first layer has stride 2 if the feature map size decreases \citep{sandler2018mobilenetv2}. All the other layers in the units have stride 1. 

We allow each unit to use arbitrary numbers of layers (denoted as \emph{elastic depth}); For each layer, we allow to use arbitrary numbers of channels (denoted as \emph{elastic width}) and arbitrary kernel sizes (denoted as \emph{elastic kernel size}). In addition, we also allow the CNN model to take arbitrary input image sizes (denoted as \emph{elastic resolution}). For example, in our experiments, the input image size ranges from 128 to 224 with a stride 4; the depth of each unit is chosen from \{2, 3, 4\}; the width expansion ratio in each layer is chosen from \{3, 4, 6\}; the kernel size is chosen from \{3, 5, 7\}. Therefore, with 5 units, we have roughly $((3\times3)^2 + (3\times3)^3 + (3\times3)^4)^5 \approx 2 \times 10^{19}$ different neural network architectures and each of them can be used under 25 different input resolutions. Since all of these sub-networks share the same weights (i.e., $W_o$) \citep{Cheung2019superposition}, we only require 7.7M parameters to store all of them. Without sharing, the total model size will be prohibitive.

\subsection{Training the \motherNetCap}
\label{sec:ofa_train}

\begin{figure}[t]
\vspace{-20pt}
    \centering
    \includegraphics[width=1\linewidth]{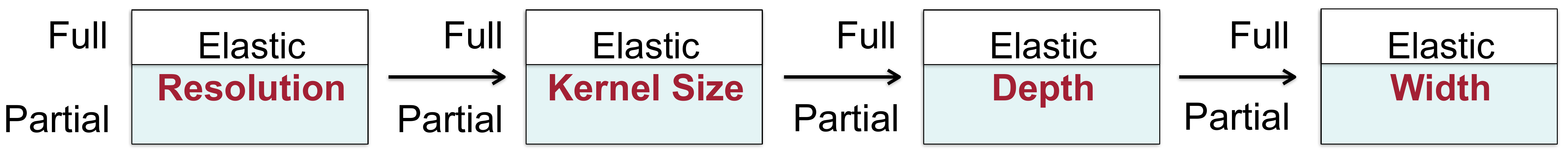}
    \caption{Illustration of the progressive shrinking process to support different depth $D$, width $W$, kernel size $K$ and resolution $R$. It leads to a large space comprising diverse sub-networks ($> 10^{19}$).}\label{fig:algo_flowchart}
\end{figure}

\begin{figure}[t]
\vspace{-10pt}
    \centering
    \includegraphics[width=1\linewidth]{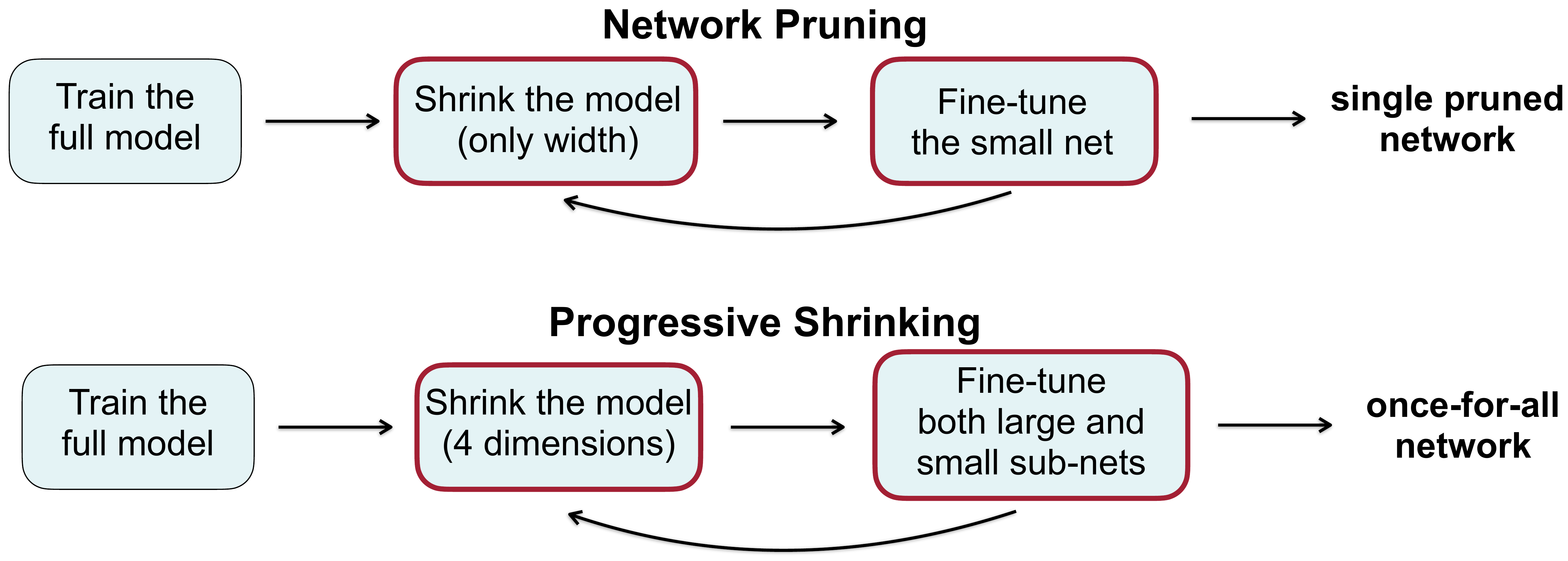}
    \caption{Progressive shrinking can be viewed as a generalized network pruning technique with much higher flexibility. Compared to network pruning, it shrinks more dimensions (not only width) and provides a much more powerful once-for-all network that can fit different deployment scenarios rather than a single pruned network.}\label{fig:ps_vs_pruning}
\end{figure}

\myparagraph{Na\"{i}ve Approach.} Training the \motherNet~can be cast as a multi-objective problem, where each objective corresponds to one sub-network. From this perspective, a na\"{i}ve training approach is to directly optimize the  \motherNet~from scratch using the exact gradient of the overall objective, which is derived by enumerating all sub-networks in each update step, as shown in Eq.~(\ref{eq:objective}). The cost of this approach is linear to the number of sub-networks. Therefore, it is only applicable to scenarios where a limited number of sub-networks are supported \citep{yu2018slimmable}, while in our case, it is computationally prohibitive to adopt this approach. 

Another na\"{i}ve training approach is to sample a few sub-networks in each update step rather than enumerate all of them, which does not have the issue of prohibitive cost. However, with such a large number of sub-networks that share weights, thus interfere with each other, we find it suffers from significant accuracy drop. In the following section, we introduce a solution to address this challenge, i.e., \emph{progressive shrinking}. 

\myparagraph{Progressive Shrinking.} The \motherNet~comprises many sub-networks of different sizes where small sub-networks are nested in large sub-networks. 
To prevent interference between the sub-networks, we propose to enforce a training order from large sub-networks to small sub-networks in a progressive manner. We name this training scheme as \emph{progressive shrinking} (PS). An example of the training process with PS is provided in Figure~\ref{fig:algo_flowchart} and Figure~\ref{fig:ps_vs_pruning}, where we start with training the largest neural network with the maximum kernel size (e.g., 7), depth (e.g., 4), and width (e.g., 6). Next, we progressively fine-tune the network to support smaller sub-networks by gradually adding them into the sampling space (larger sub-networks may also be sampled). Specifically, after training the largest network, we first support elastic kernel size which can choose from \{3, 5, 7\} at each layer, while the depth and width remain the maximum values. Then, we support elastic depth and elastic width sequentially, as is shown in Figure~\ref{fig:algo_flowchart}. The resolution is elastic throughout the whole training process, which is implemented by sampling different image sizes for each batch of training data. We also use the knowledge distillation technique after training the largest neural network \citep{hinton2015distilling,ashok2017n2n,yu2019universally}. It combines two loss terms using both the soft labels given by the largest neural network and the real labels.

Compared to the na\"{i}ve approach, PS prevents small sub-networks from interfering large sub-networks, since large sub-networks are already well-trained when the \motherNet~is fine-tuned to support small sub-networks.  Regarding the small sub-networks, they share the weights with the large ones. Therefore, PS allows initializing small sub-networks with the most important weights of well-trained large sub-networks, which expedites the training process. Compared to network pruning (Figure~\ref{fig:ps_vs_pruning}), PS also starts with training the full model, but it shrinks not only the width dimension but also the depth, kernel size, and resolution dimensions of the full model. Additionally, PS fine-tunes both large and small sub-networks rather than a single pruned network. As a result, PS provides a much more powerful once-for-all network that can fit diverse hardware platforms and efficiency constraints compared to network pruning. 
We describe the details of the PS training flow as follows:

\begin{figure}[t]
    \centering
    \includegraphics[width=1.0\linewidth]{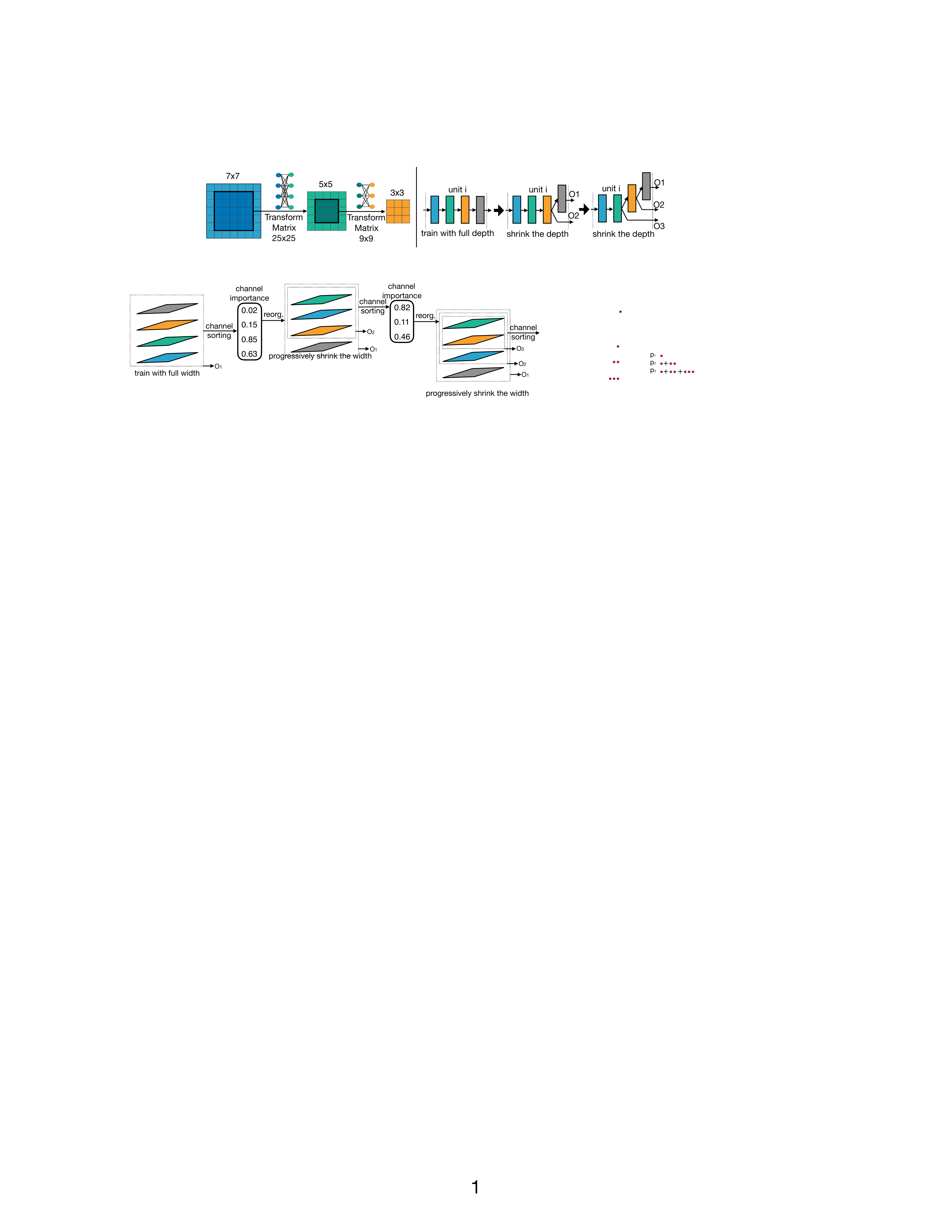}
    \caption{Left: Kernel transformation matrix for elastic kernel size. Right: Progressive shrinking for elastic depth. Instead of skipping each layer independently, we keep the first $D$ layers and skip the last $(4 - D)$ layers. The weights of the early layers are shared.}\label{fig:elastic_ks_depth}
\end{figure}

\begin{figure}[t]
    \centering
    \includegraphics[width=1\linewidth]{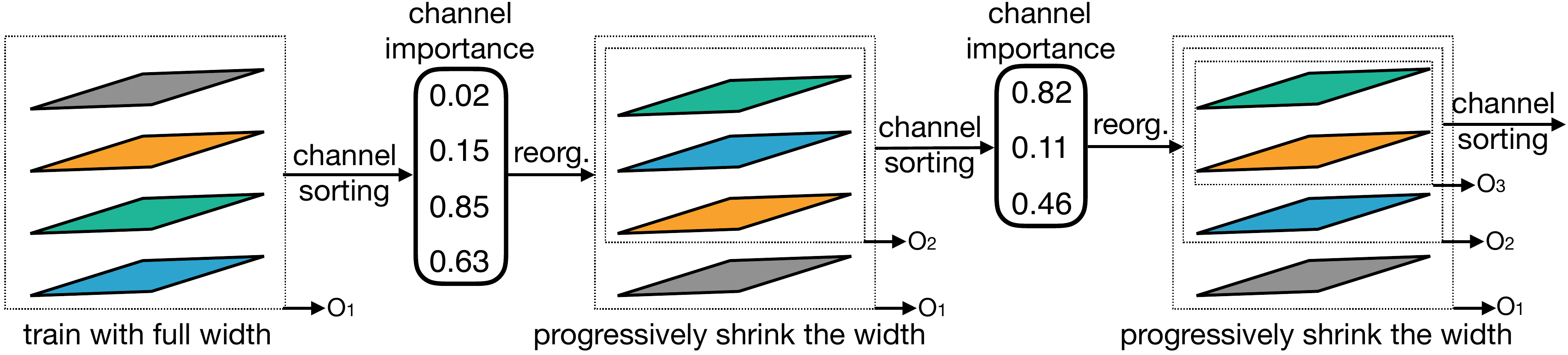}
    \caption{Progressive shrinking for elastic width. In this example, we progressively support 4, 3, and 2 channel settings. We perform channel sorting and pick the most important channels (with large L1 norm) to initialize the smaller channel settings. The important channels' weights are shared.}\label{fig:elastic_width}
\end{figure}

\begin{itemize}[leftmargin=*]

\item \textbf{Elastic Kernel Size} (Figure~\ref{fig:elastic_ks_depth} left). We let the center of a 7x7 convolution kernel also serve as a 5x5 kernel, the center of which can also be a 3x3 kernel. Therefore, the kernel size becomes elastic.
The challenge is that the centering sub-kernels (e.g., 3x3 and 5x5) are shared and need to play multiple roles (independent kernel and part of a large kernel). The weights of centered sub-kernels may need to have different distribution or magnitude as different roles. Forcing them to be the same degrades the performance of some sub-networks. Therefore, we introduce kernel transformation matrices when sharing the kernel weights. We use separate kernel transformation matrices for different layers. Within each layer, the kernel transformation matrices are shared among different channels. As such, we only need $25 \times 25 + 9 \times 9 = 706$ extra parameters to store the kernel transformation matrices in each layer, which is negligible. 

\item \textbf{Elastic Depth}  (Figure~\ref{fig:elastic_ks_depth} right). 
To derive a sub-network that has $D$ layers in a unit that originally has $N$ layers, we keep the \textit{first} D layers and skip the last $N-D$ layers, rather than keeping \textit{any} $D$ layers as done in current NAS methods \citep{cai2019proxylessnas,wu2018fbnet}. As such, one depth setting only corresponds to one combination of layers. In the end, the weights of the first D layers are shared between large and small models.

\item \textbf{Elastic Width} (Figure~\ref{fig:elastic_width}). Width means the number of channels. We give each layer the flexibility to choose different channel expansion ratios. Following the progressive shrinking scheme, we first train a full-width model. 
Then we introduce a channel sorting operation to support partial widths. It reorganizes the channels according to their importance, which is calculated based on the L1 norm of a channel's weight. Larger L1 norm means more important. For example, when shrinking from a 4-channel-layer to a 3-channel-layer, we select the largest 3 channels, whose weights are shared with the 4-channel-layer (Figure~\ref{fig:elastic_width} left and middle). Thereby, smaller sub-networks are initialized with the most important channels on the \motherNet~which is already well trained. This channel sorting operation preserves the accuracy of larger sub-networks.  

\end{itemize}

\subsection{Specialized Model Deployment with~\motherNetCap}
\label{sec:ofa_search}
Having trained a \motherNet, the next stage is to derive the specialized sub-network for a given deployment scenario. The goal is to search for a neural network that satisfies the efficiency (e.g., latency, energy) constraints on the target hardware while optimizing the accuracy. Since OFA decouples model training from neural architecture search, we do not need any training cost in this stage. Furthermore, we build \emph{neural-network-twins} to predict the latency and accuracy given a neural network architecture, providing a quick feedback for model quality. It eliminates the repeated search cost by substituting the measured accuracy/latency with predicted accuracy/latency (twins).

Specifically, we randomly sample 16K sub-networks with different architectures and input image sizes, then measure their accuracy on 10K validation images sampled from the original training set. These [architecture, accuracy] pairs are used to train an accuracy predictor to predict the accuracy of a model given its architecture and input image size\footnote{Details of the accuracy predictor is provided in Appendix~\ref{appendix:acc_predictor}.}. We also build a latency lookup table \citep{cai2019proxylessnas} on each target hardware platform to predict the latency. Given the target hardware and latency constraint, we conduct an evolutionary search \citep{real2018regularized} based on the neural-network-twins to get a specialized sub-network. 
Since the cost of searching with neural-network-twins is negligible, we only need 40 GPU hours to collect the data pairs, and the cost stays constant regardless of \#deployment scenarios.

\section{Experiments}

\begin{figure}[t]
\vspace{-25pt}
    \centering
    \includegraphics[width=0.9\linewidth]{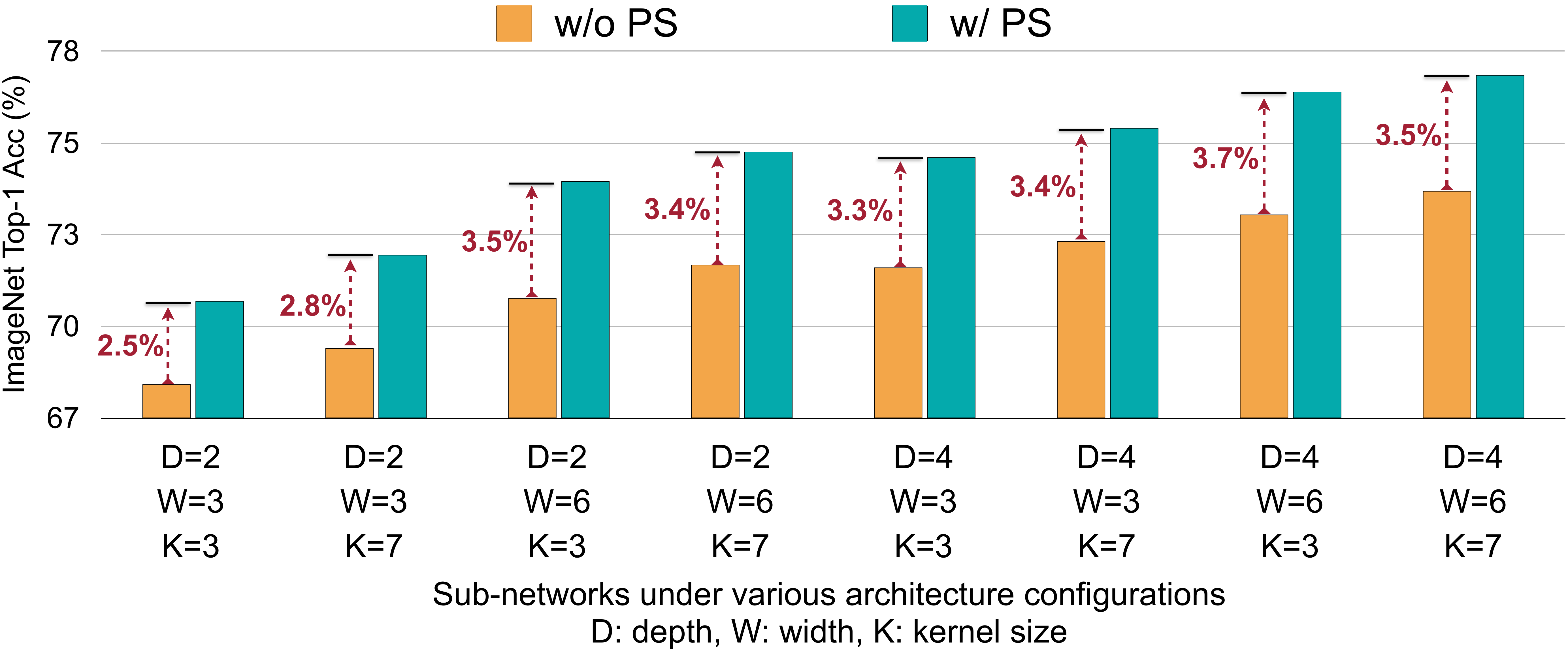}
    \caption{ImageNet top1 accuracy (\%) performances of sub-networks under resolution $224\times224$. ``(D = $d$, W = $w$, K = $k$)'' denotes a sub-network with $d$ layers in each unit, and each layer has an width expansion ratio $w$ and kernel size $k$.}\label{fig:paranet_net_on_imagenet}
\end{figure}

In this section, we first apply the progressive shrinking algorithm to train the \motherNet~on ImageNet \citep{deng2009imagenet}. Then we demonstrate the effectiveness of our trained \motherNet~on various hardware platforms (Samsung S7 Edge, Note8, Note10, Google Pixel1, Pixel2, LG G8, NVIDIA 1080Ti, V100 GPUs, Jetson TX2, Intel Xeon CPU, Xilinx ZU9EG, and ZU3EG FPGAs) with different latency constraints. 

\subsection{Training the \motherNetCap~on ImageNet}

\myparagraph{Training Details.} We use the same architecture space as MobileNetV3 \citep{howard2019searching}. For training the full network, we use the standard SGD optimizer with Nesterov momentum 0.9 and weight decay $3e^{-5}$. The initial learning rate is 2.6, and we use the cosine schedule \citep{loshchilov2016sgdr} for learning rate decay. The full network is trained for 180 epochs with batch size 2048 on 32 GPUs. Then we follow the schedule described in Figure~\ref{fig:algo_flowchart} to further fine-tune the full network\footnote{Implementation details can be found in Appendix~\ref{appendix:ps_details}.}. The whole training process takes around 1,200 GPU hours on V100 GPUs. This is a one-time training cost that can be amortized by many deployment scenarios. 

\myparagraph{Results.} Figure~\ref{fig:paranet_net_on_imagenet} reports the top1 accuracy of sub-networks derived from the \motherNet s that are trained with our progressive shrinking (PS) algorithm and without PS respectively. Due to space limits, we take 8 sub-networks for comparison, and each of them is denoted as ``(D = $d$, W = $w$, K = $k$)''. It represents a sub-network that has $d$ layers for all units, while the expansion ratio and kernel size are set to $w$ and $k$ for all layers. PS can improve the ImageNet accuracy of sub-networks by a significant margin under all architectural settings. Specifically, without architecture optimization, PS can achieve 74.8\% top1 accuracy using 226M MACs under the architecture setting (D=4, W=3, K=3), which is on par with MobileNetV3-Large. In contrast, without PS, it only achieves 71.5\%, which is 3.3\% lower.

\subsection{Specialized \childNetCap s for Different Hardware and Constraints}

We apply our trained \motherNet~to get different specialized sub-networks for diverse hardware platforms: from the cloud to the edge. \textbf{On cloud devices}, the latency for GPU is measured with batch size 64 on NVIDIA 1080Ti and V100 with Pytorch 1.0+cuDNN. The CPU latency is measured with batch size 1 on Intel Xeon E5-2690 v4+MKL-DNN. \textbf{On edge devices}, including mobile phones, we use Samsung, Google and LG phones with TF-Lite, batch size 1; for mobile GPU, we use Jetson TX2 with Pytorch 1.0+cuDNN, batch size of 16; for embedded FPGA, we use Xilinx ZU9EG and ZU3EG FPGAs with Vitis AI\footnote{https://www.xilinx.com/products/design-tools/vitis/vitis-ai.html}, batch size 1. 

\begin{table}[!t]
\centering
\vspace{-20pt}
\resizebox{1\linewidth}{!}{
\begin{tabular}{l | c | c | c | c | c | c | c | c }
    \hline
    \multirow{2}{*}{Model} & ImageNet & \multirow{2}{*}{MACs} & Mobile & Search cost & Training cost & \multicolumn{3}{c}{Total cost $(N = 40)$} \\
     \cline{7-9}
    & Top1 (\%) & & latency & (GPU hours) & (GPU hours) & GPU hours & $CO_2$e (lbs) & AWS cost \\
    \hline
    MobileNetV2 [\citenum{sandler2018mobilenetv2}] & 72.0 & 300M & 66ms & 0 & 150$N$ & 6k & 1.7k & \$18.4k \\
    MobileNetV2 \#1200 & 73.5 & 300M & 66ms & 0 & 1200$N$ & 48k & 13.6k & \$146.9k  \\
    \hline 
    NASNet-A [\citenum{zoph2018learning}] & 74.0 & 564M & - & 48,000$N$ & - & 1,920k & 544.5k & \$5875.2k  \\
    DARTS [\citenum{liu2019darts}] & 73.1 & 595M & - & 96$N$ & 250$N$ & 14k & 4.0k & \$42.8k \\
    \hline
    MnasNet [\citenum{tan2018mnasnet}] & 74.0 & 317M & 70ms & 40,000$N$ & - & 1,600k & 453.8k & \$4896.0k \\
    FBNet-C [\citenum{wu2018fbnet}] & 74.9 & 375M & - & 216$N$ & 360$N$ & 23k & 6.5k & \$70.4k \\
    ProxylessNAS [\citenum{cai2019proxylessnas}] & 74.6 & 320M & 71ms & 200$N$ & 300$N$ & 20k & 5.7k & \$61.2k \\
    SinglePathNAS [\citenum{guo2019single}] & 74.7 & 328M & - & 288 + 24$N$ & 384$N$ & 17k & 4.8k & \$52.0k \\
    AutoSlim [\citenum{yu2019autoslim}] & 74.2 & 305M & 63ms & 180 & 300$N$ & 12k & 3.4k & \$36.7k \\
    MobileNetV3-Large [\citenum{howard2019searching}] & 75.2 & 219M & 58ms & - & 180$N$ & 7.2k & 1.8k & \$22.2k \\
    \hline
    OFA w/o PS & 72.4 & 235M & 59ms & 40 & 1200 & 1.2k & 0.34k & \$3.7k \\
    OFA w/ PS & \textbf{76.0} & 230M & 58ms & 40 & 1200 & 1.2k  & 0.34k & \$3.7k \\
    OFA w/ PS \#25 & \textbf{76.4} & 230M & 58ms & 40 & 1200 + 25$N$ & 2.2k & 0.62k & \$6.7k \\
    OFA w/ PS \#75 & \textbf{76.9} & 230M & 58ms & 40 & 1200 + 75$N$ & 4.2k & 1.2k & \$13.0k \\
    \hline
    OFA$_{\text{Large}}$ w/ PS \#75 & \textbf{80.0} & 595M & - & 40 & 1200 + 75$N$ & 4.2k & 1.2k & \$13.0k \\
    \hline
\end{tabular}
}
\caption{Comparison with SOTA hardware-aware NAS methods on Pixel1 phone.
OFA decouples model training from neural architecture search. The search cost and training cost both stay constant as the number of deployment scenarios grows. ``\#25'' denotes the specialized sub-networks are fine-tuned for 25 epochs after grabbing weights from the \motherNet. ``$CO_2e$'' denotes $CO_2$ emission which is calculated based on \cite{strubell2019energy}. AWS cost is calculated based on the price of on-demand P3.16xlarge instances.
}
\label{tab:comparison_with_nas}
\end{table}

\begin{figure}[t]
    \centering
    \includegraphics[width=0.7\linewidth]{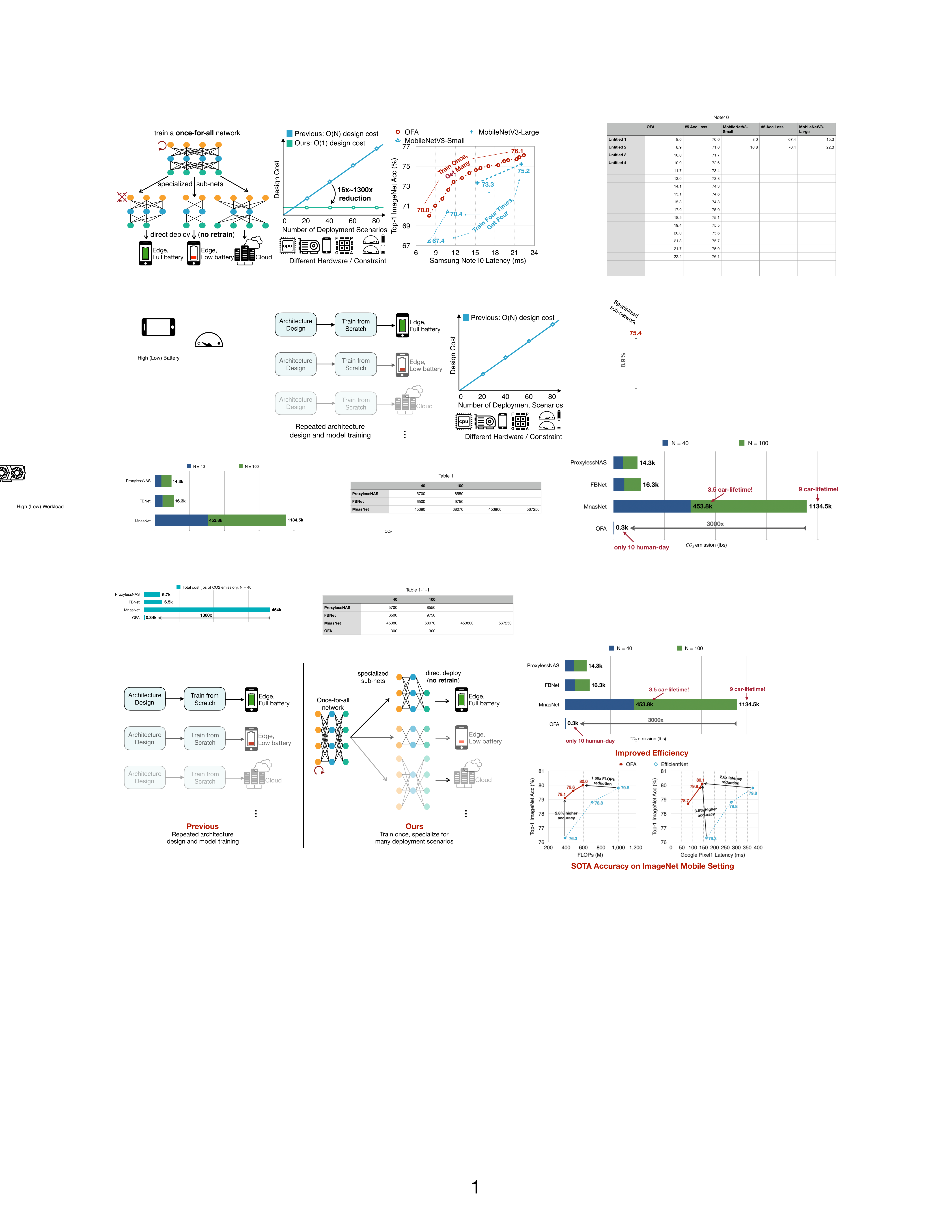}
    \caption{OFA saves orders of magnitude design cost compared to NAS methods.}\label{fig:search_cost_bar}
\end{figure}

\myparagraph{Comparison with NAS on Mobile Devices.}
Table~\ref{tab:comparison_with_nas} reports the comparison between OFA and state-of-the-art hardware-aware NAS methods on the mobile phone (Pixel1). 
OFA is much more efficient than NAS when handling multiple deployment scenarios since the cost of OFA is \textit{constant} while others are \textit{linear} to the number of deployment scenarios ($N$). \textbf{With $N$ = 40, the total $CO_2$ emissions of OFA is 16$\times$ fewer than ProxylessNAS, 19$\times$ fewer than FBNet, and 1,300$\times$ fewer than MnasNet (Figure~\ref{fig:search_cost_bar}).}
Without retraining, OFA achieves 76.0\% top1 accuracy on ImageNet, which is 0.8\% higher than MobileNetV3-Large while maintaining similar mobile latency.
We can further improve the top1 accuracy to 76.4\% by fine-tuning the specialized sub-network for 25 epochs and to 76.9\% by fine-tuning for 75 epochs. Besides, we also observe that OFA with PS can achieve 3.6\% better accuracy than without PS. 

\begin{figure}[t]
    \centering
    \vspace{-20pt}
    \includegraphics[width=0.9\textwidth]{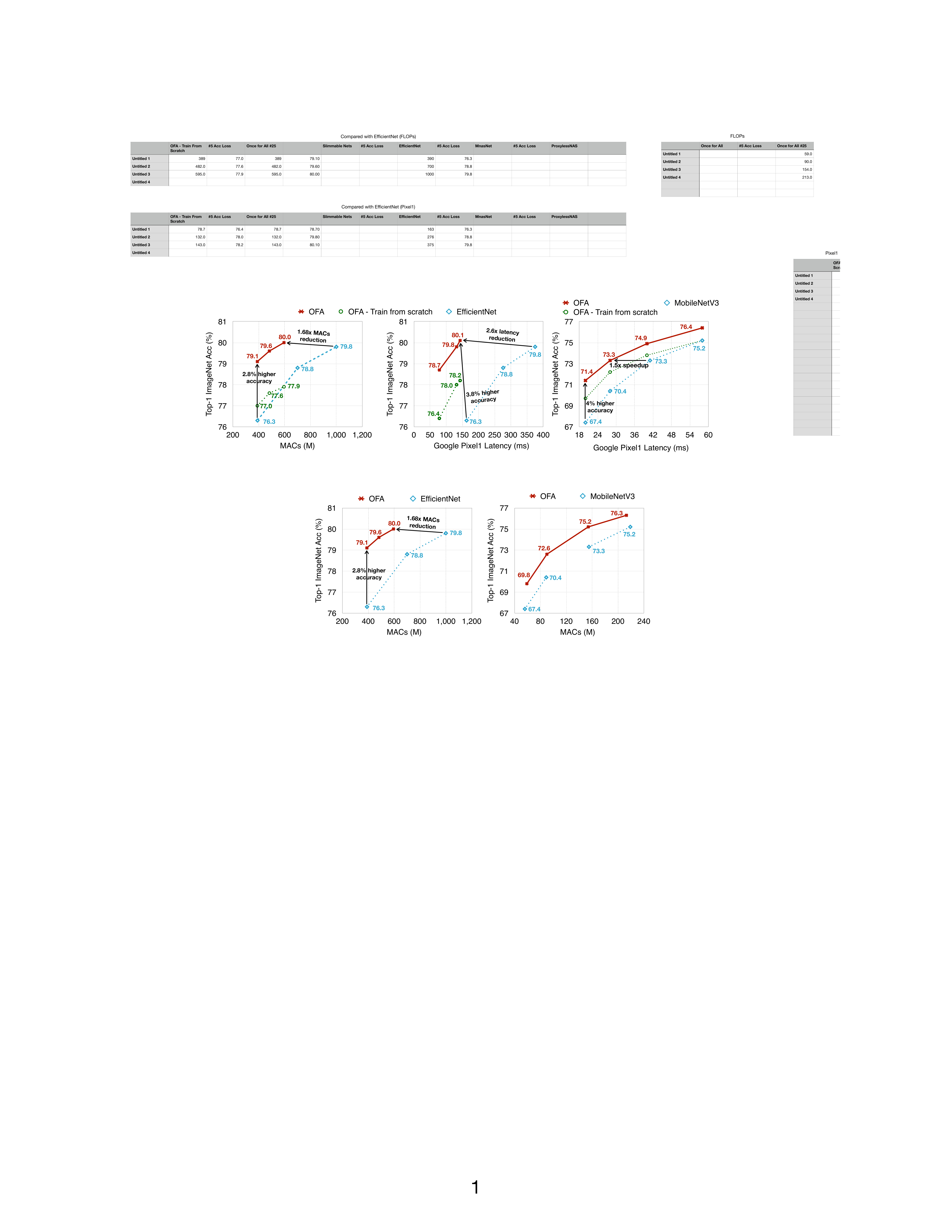}
    \caption{OFA achieves 80.0\% top1 accuracy with 595M MACs and 80.1\% top1 accuracy with 143ms Pixel1 latency, setting a new SOTA ImageNet top1 accuracy on the mobile setting.}
    \label{fig:mobile_80}
\end{figure}

\myparagraph{OFA under Different Computational Resource Constraints.} Figure~\ref{fig:mobile_80} summarizes the results of OFA under different MACs and Pixel1 latency constraints. OFA achieves 79.1\% ImageNet top1 accuracy with 389M MACs, being 2.8\% more accurate than EfficientNet-B0 that has similar MACs. With 595M MACs, OFA reaches a new SOTA 80.0\% ImageNet top1 accuracy under the mobile setting ($<$600M MACs), which is 0.2\% higher than EfficientNet-B2 while using 1.68$\times$ fewer MACs. More importantly, OFA runs much faster than EfficientNets on hardware. Specifically, with 143ms Pixel1 latency, OFA achieves 80.1\% ImageNet top1 accuracy, being 0.3\% more accurate and 2.6$\times$ faster than EfficientNet-B2. We also find that training the searched neural architectures from scratch cannot reach the same level of accuracy as OFA, suggesting that not only neural architectures but also pre-trained weights contribute to the superior performances of OFA. 

\begin{figure}[t]
    \centering
    \includegraphics[width=1\textwidth]{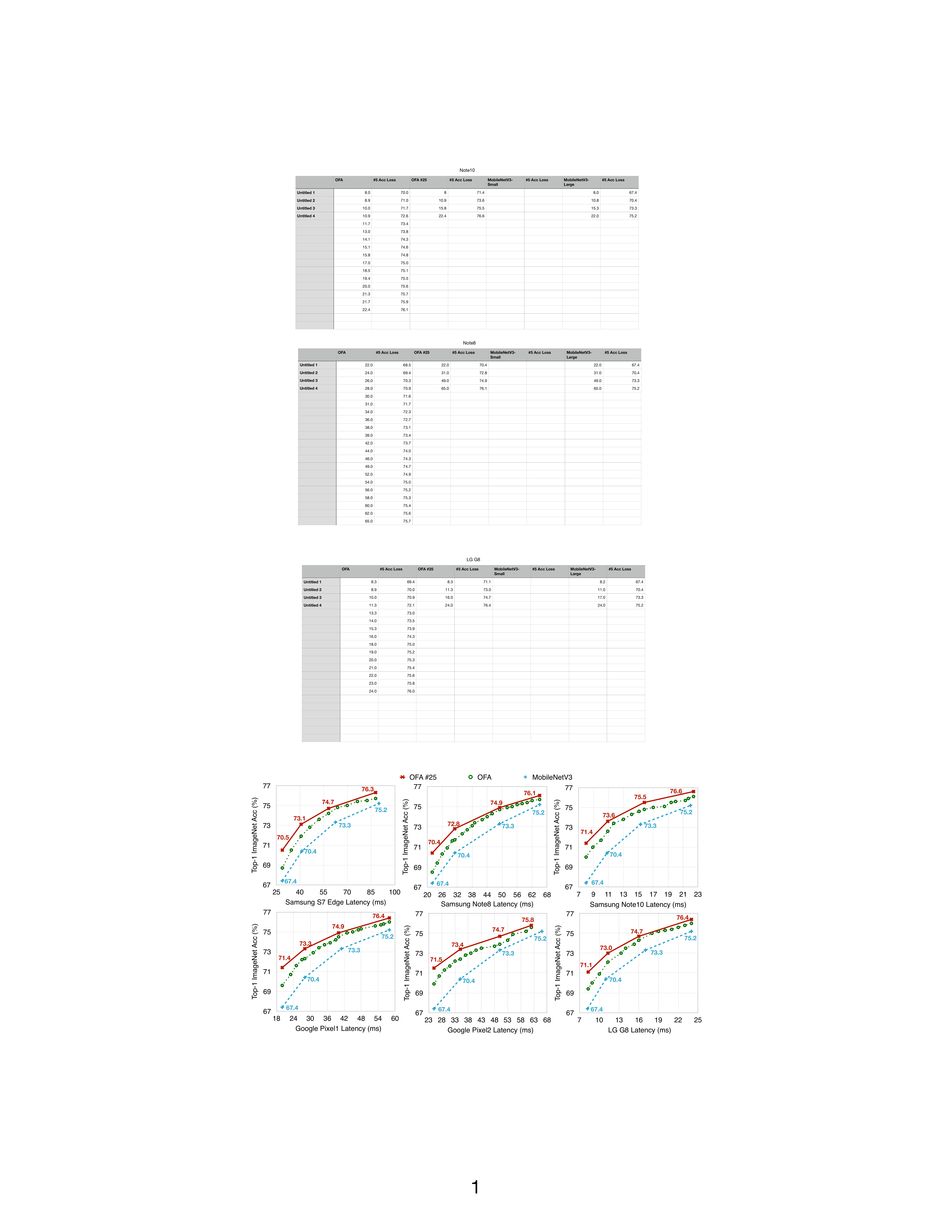}
    \caption{OFA consistently outperforms MobileNetV3 on mobile platforms.}
    \label{fig:new_mobile}
\end{figure}

Figure~\ref{fig:new_mobile} reports detailed comparisons between OFA and MobileNetV3 on six mobile devices. Remarkably, \textbf{OFA can produce the entire trade-off curves with many points over a wide range of latency constraints by training only once} (green curve). It is impossible for previous NAS methods \citep{tan2018mnasnet,cai2019proxylessnas} due to the prohibitive training cost. 

\begin{figure}[t]
    \centering
    \vspace{-25pt}
    \includegraphics[width=\linewidth]{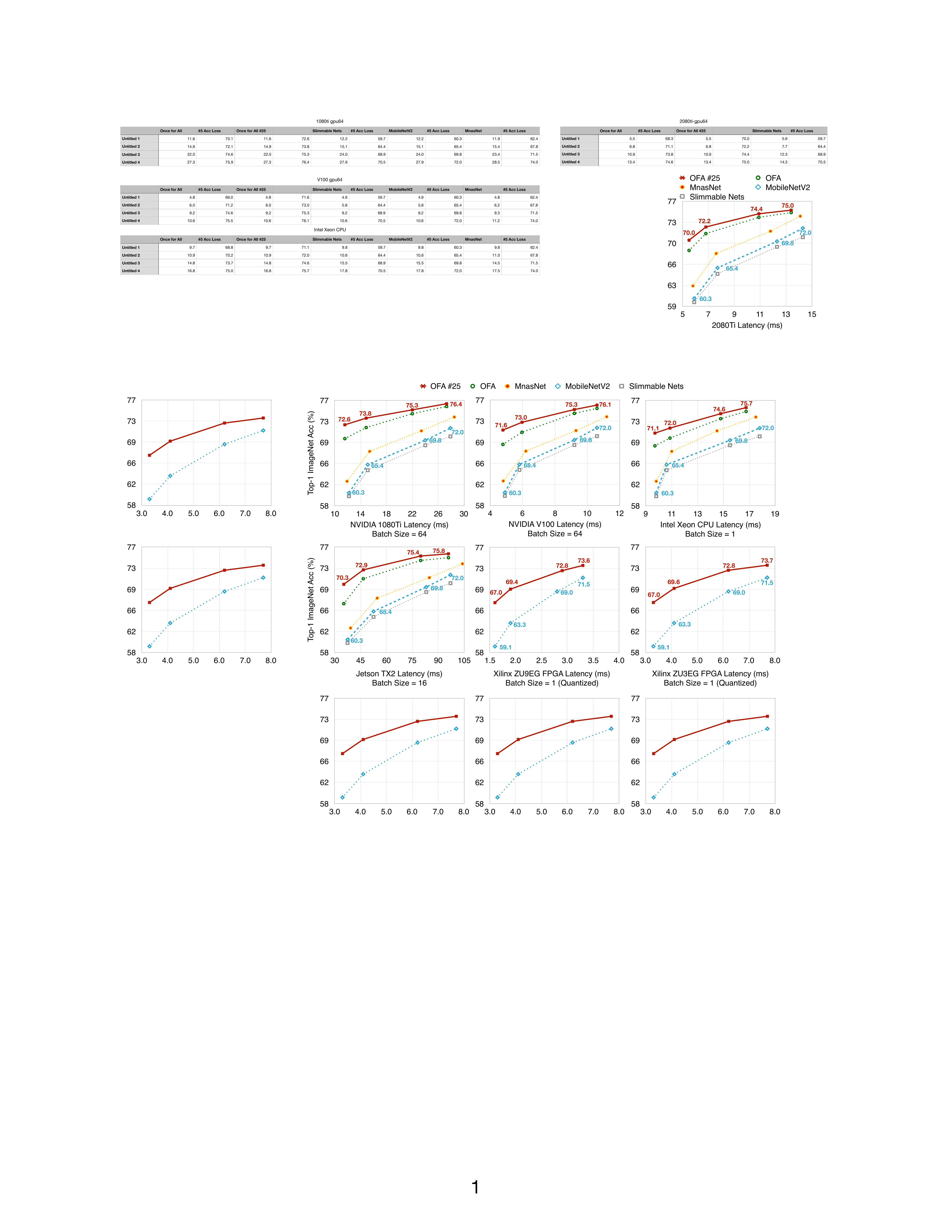}
    \caption{
        Specialized OFA models consistently achieve significantly higher ImageNet accuracy with similar latency than non-specialized neural networks on CPU, GPU, mGPU, and FPGA. More remarkably, specializing for a new hardware platform does not add training cost using OFA.
    }
    \label{fig:specialized_different_hardwares}
\end{figure}

\begin{figure}[ht]
    \centering
    \begin{tabular}{l}
        \begin{minipage}{0.75\linewidth}
            \includegraphics[width=1\linewidth]{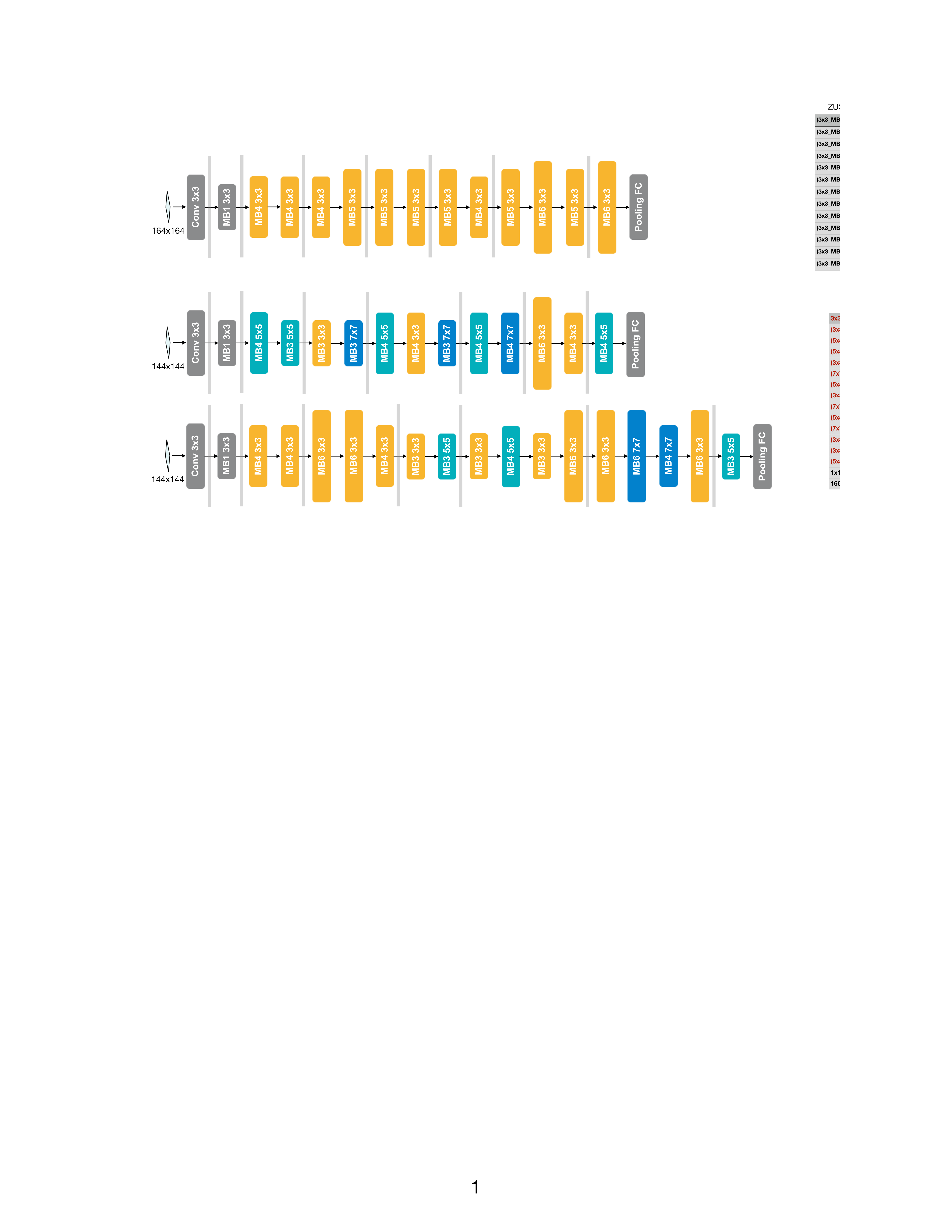}
        \end{minipage}
        \\
        \multicolumn{1}{c}{ (a) 4.1ms latency on Xilinx ZU3EG (batch size = 1).}
        \vspace{6pt}
        \\
        \begin{minipage}{0.75\linewidth}
            \includegraphics[width=\linewidth]{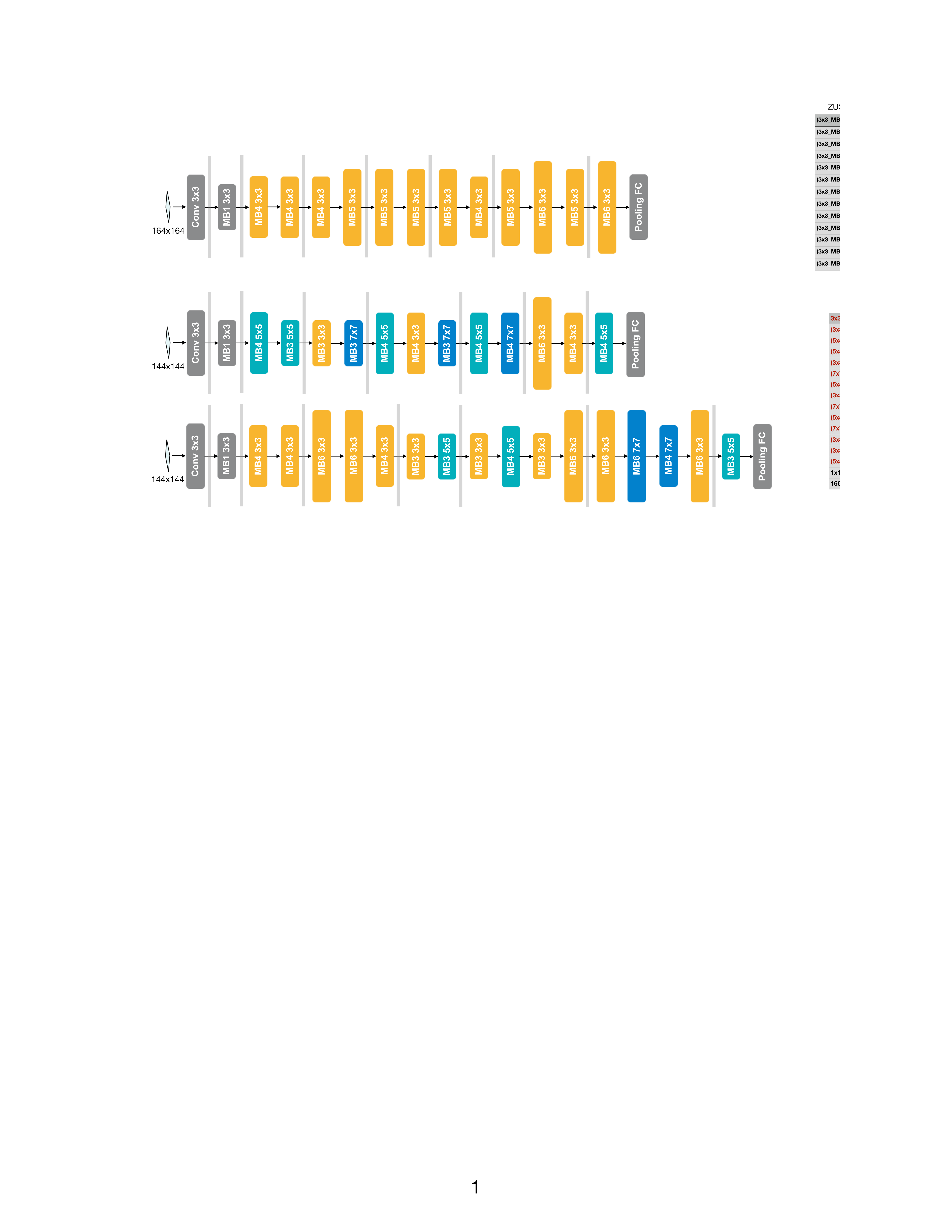}
        \end{minipage}
        \\
        \multicolumn{1}{c}{ (b) 10.9ms latency on Intel Xeon CPU (batch size = 1).}
        \vspace{6pt}
        \\
        \begin{minipage}{0.999\linewidth}
            \includegraphics[width=1\linewidth]{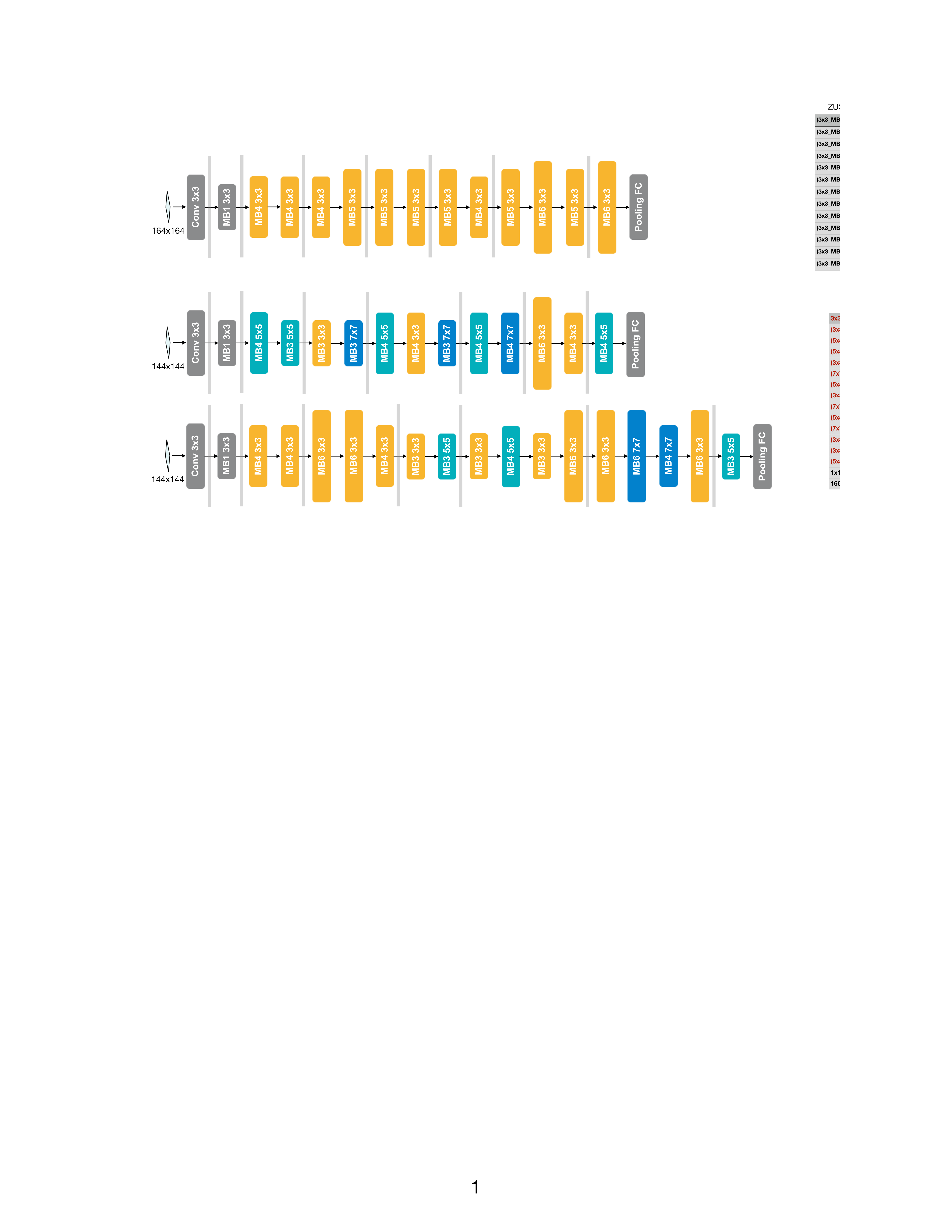}
        \end{minipage}
        \\
        \multicolumn{1}{c}{ (c) 14.9ms latency on NVIDIA 1080Ti (batch size = 64).}
    \end{tabular}
    \caption{OFA can design specialized models for different hardware and different latency constraint. ``MB4 3x3'' means ``mobile block with expansion ratio 4, kernel size 3x3''. FPGA and GPU models are wider than CPU model due to larger parallelism. Different hardware has different cost model, leading to different optimal CNN architectures. OFA provides a unified and efficient design methodology.}
    \label{fig:model_architectures}
\end{figure}

\myparagraph{OFA for Diverse Hardware Platforms.}
Besides the mobile platforms, we extensively studied the effectiveness of OFA on six additional hardware platforms (Figure~\ref{fig:specialized_different_hardwares}) using the ProxylessNAS architecture space \citep{cai2019proxylessnas}.
OFA consistently improves the trade-off between accuracy and latency by a significant margin, especially on GPUs which have more parallelism. With similar latency as MobileNetV2 0.35, ``OFA \#25'' improves the ImageNet top1 accuracy from MobileNetV2's 60.3\% to 72.6\% (+12.3\% improvement) on the 1080Ti GPU. Detailed architectures of our specialized models are shown in Figure~\ref{fig:model_architectures}. 
It reveals the insight that using the \textit{same} model for different deployment scenarios with \textit{only} the width multiplier modified has a limited impact on efficiency improvement: the accuracy drops quickly as the latency constraint gets tighter.

\begin{figure}[t]
    \centering
    \includegraphics[width=\linewidth]{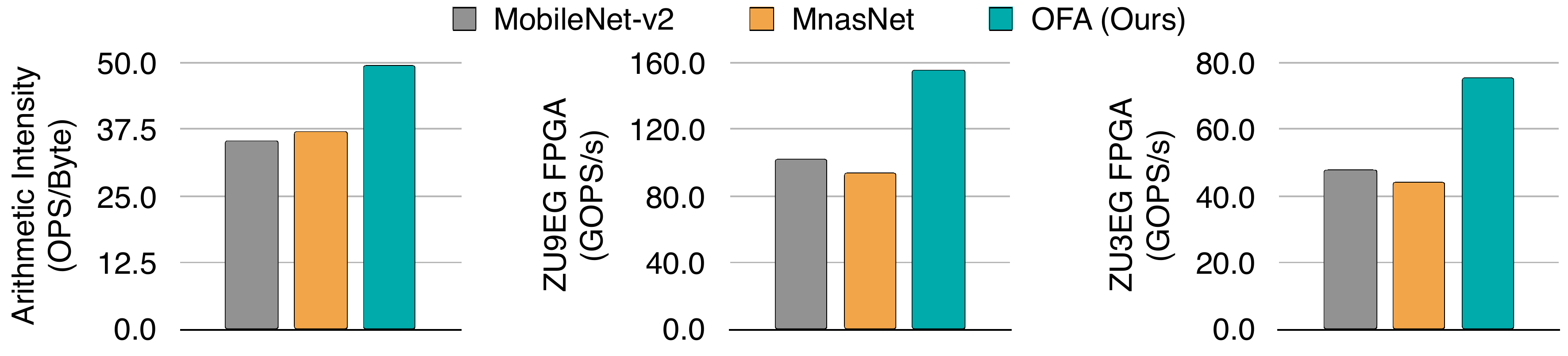}
    \caption{OFA models improve the arithmetic intensity (OPS/Byte) and utilization (GOPS/s) compared with the MobileNetV2 and MnasNet (measured results on Xilinx ZU9EG and ZU3EG FPGA). 
    }
    \label{fig:fpga_numbers}
\end{figure}

\myparagraph{OFA for Specialized Hardware Accelerators.}
There has been plenty of work on NAS for general-purpose hardware, but little work has been focused on specialized hardware accelerators. We quantitatively analyzed the performance of OFA on two FPGAs accelerators (ZU3EG and ZU9EG) using Xilinx Vitis AI with 8-bit quantization, and discuss two design principles.  

\begin{figure}[t]
    \vspace{-25pt}
    \centering
    \includegraphics[width=\linewidth]{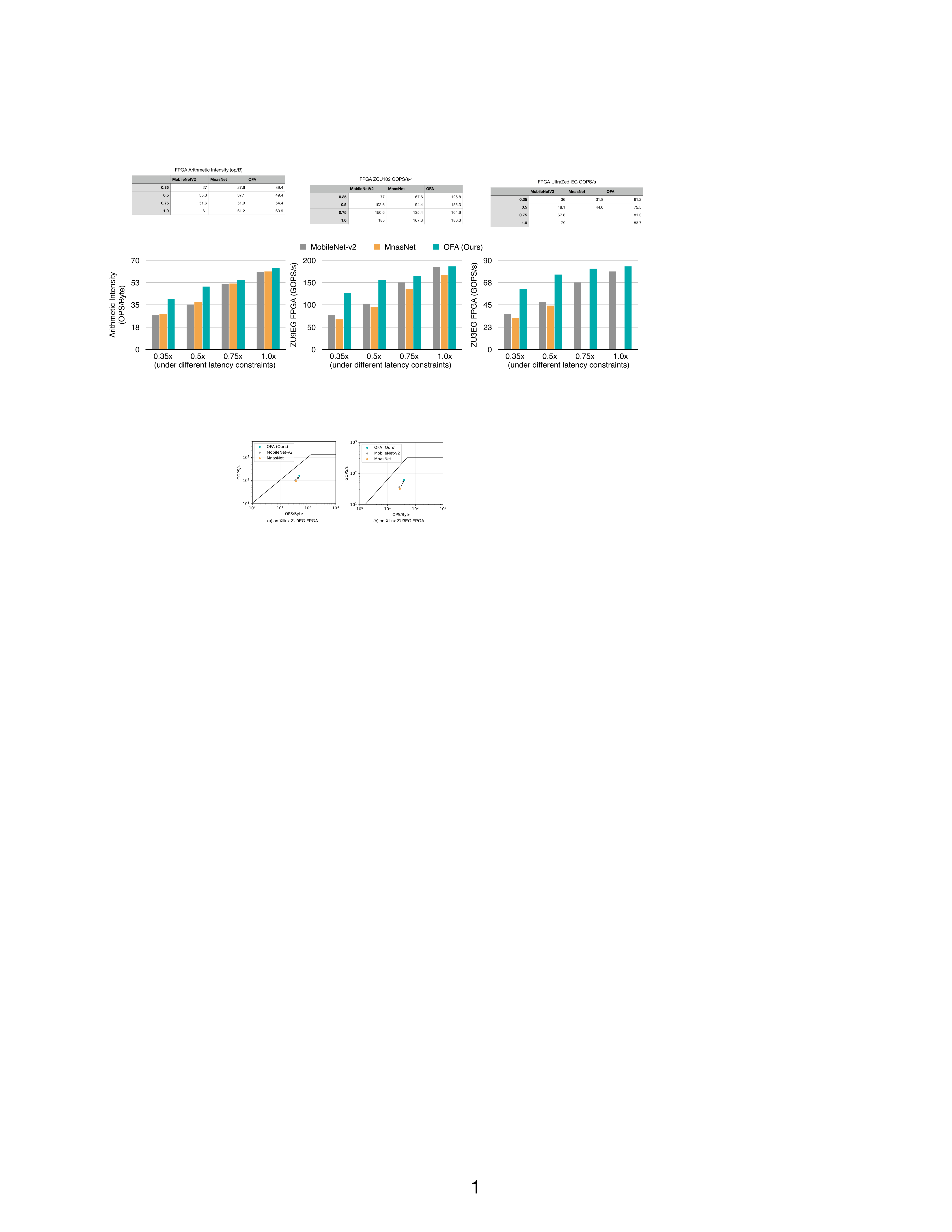}
    \caption{Quantative study of OFA's roofline model on Xilinx ZU9EG and ZU3EG FPGAs (log scale). OFA model increased the arithmetic intensity by 33$\%$/43$\%$ and GOPS/s by 72$\%$/92$\%$ on these two FPGAs compared with MnasNet. 
    }
    \vspace{-5pt}
    \label{fig:fpga_roofline}
\end{figure}

\textbf{Principle 1}: memory access is expensive, computation is cheap. An efficient CNN should do \textit{as much as} computation with \textit{a small amount} of memory footprint. The ratio is defined as the arithmetic intensity (OPs/Byte). The higher OPs/Byte, the less memory bounded, the easier to parallelize. Thanks to OFA's diverse choices of sub-network architectures ($10^{19}$) (Section~\ref{sec:ofa_train}), and the OFA model twin that can quickly give the accuracy/latency feedback (Section~\ref{sec:ofa_search}), the evolutionary search can automatically find a CNN architecture that has higher arithmetic intensity. As shown in Figure~\ref{fig:fpga_numbers}, OFA's arithmetic intensity is 48\%/43\% higher than MobileNetV2 and MnasNet (MobileNetV3 is not supported by Xilinx Vitis AI). Removing the memory bottleneck results in higher utilization and GOPS/s by 70\%-90\%, pushing the operation point to the upper-right in the roofline model~\citep{williams2009roofline}, as shown in Figure~\ref{fig:fpga_roofline}. (70\%-90\% looks small in the log scale but that is significant). 

\textbf{Principle 2}: the CNN architecture should be co-designed with the hardware accelerator's cost model. The FPGA accelerator has a specialized depth-wise engine that is pipelined with the point-wise engine. The pipeline throughput is perfectly matched for 3x3 kernels. As a result, OFA's searched model only has 3x3 kernel (Figure~\ref{fig:model_architectures}, a) on FPGA, despite 5x5 and 7x7 kernels are also in the search space. Additionally, large kernels sometimes cause ``out of BRAM'' error on FPGA, giving high cost. On Intel Xeon CPU, however, more than 50\% operations are large kernels. Both FPGA and GPU models are wider than CPU, due to the large parallelism of the computation array. 


\section{Conclusion}
We proposed \textit{Once-for-All} (OFA), a new methodology that decouples model training from architecture search for efficient deep learning deployment under a large number of hardware platforms. Unlike previous approaches that design and train a neural network for \textit{each} deployment scenario, we designed a \textit{\motherNet}~that supports different architectural configurations, including elastic depth, width, kernel size, and resolution. It reduces the training cost (GPU hours, energy consumption, and $CO_2$ emission) by orders of magnitude compared to conventional methods. 
To prevent sub-networks of different sizes from interference, we proposed a progressive shrinking algorithm that enables a large number of \childNet~to achieve the same level of accuracy compared to training them independently. 
Experiments on a diverse range of hardware platforms and efficiency constraints demonstrated the effectiveness of our approach. OFA provides an automated ecosystem to efficiently design efficient neural networks with the hardware cost model in the loop.

\subsubsection*{Acknowledgments}
We thank  NSF Career Award \#1943349, MIT-IBM Watson AI Lab, Google-Daydream Research Award, Samsung, Intel, Xilinx, SONY, AWS Machine Learning Research Award  for supporting this research. We thank Samsung, Google and LG for donating mobile phones. We thank Shuang Wu and Lei Deng for drawing the Figure~\ref{fig:cnn_imagenet}.

\bibliography{ref}

\begin{thebibliography}{44}
\providecommand{\natexlab}[1]{#1}
\providecommand{\url}[1]{\texttt{#1}}
\expandafter\ifx\csname urlstyle\endcsname\relax
  \providecommand{\doi}[1]{doi: #1}\else
  \providecommand{\doi}{doi: \begingroup \urlstyle{rm}\Url}\fi

\bibitem[Ashok et~al.(2018)Ashok, Rhinehart, Beainy, and Kitani]{ashok2017n2n}
Anubhav Ashok, Nicholas Rhinehart, Fares Beainy, and Kris~M Kitani.
\newblock N2n learning: Network to network compression via policy gradient
  reinforcement learning.
\newblock In \emph{ICLR}, 2018.

\bibitem[Cai et~al.(2018{\natexlab{a}})Cai, Chen, Zhang, Yu, and
  Wang]{cai2018efficient}
Han Cai, Tianyao Chen, Weinan Zhang, Yong Yu, and Jun Wang.
\newblock Efficient architecture search by network transformation.
\newblock In \emph{AAAI}, 2018{\natexlab{a}}.

\bibitem[Cai et~al.(2018{\natexlab{b}})Cai, Yang, Zhang, Han, and
  Yu]{cai2018path}
Han Cai, Jiacheng Yang, Weinan Zhang, Song Han, and Yong Yu.
\newblock Path-level network transformation for efficient architecture search.
\newblock In \emph{ICML}, 2018{\natexlab{b}}.

\bibitem[Cai et~al.(2019)Cai, Zhu, and Han]{cai2019proxylessnas}
Han Cai, Ligeng Zhu, and Song Han.
\newblock Proxyless{NAS}: Direct neural architecture search on target task and
  hardware.
\newblock In \emph{ICLR}, 2019.
\newblock URL \url{https://arxiv.org/pdf/1812.00332.pdf}.

\bibitem[Cheung et~al.(2019)Cheung, Terekhov, Chen, Agrawal, and
  Olshausen]{Cheung2019superposition}
Brian Cheung, Alex Terekhov, Yubei Chen, Pulkit Agrawal, and Bruno Olshausen.
\newblock Superposition of many models into one.
\newblock In \emph{NeurIPS}, 2019.

\bibitem[Courbariaux et~al.(2015)Courbariaux, Bengio, and
  David]{courbariaux2015binaryconnect}
Matthieu Courbariaux, Yoshua Bengio, and Jean-Pierre David.
\newblock Binaryconnect: Training deep neural networks with binary weights
  during propagations.
\newblock In \emph{NeurIPS}, 2015.

\bibitem[Deng et~al.(2009)Deng, Dong, Socher, Li, Li, and
  Fei-Fei]{deng2009imagenet}
Jia Deng, Wei Dong, Richard Socher, Li-Jia Li, Kai Li, and Li~Fei-Fei.
\newblock Imagenet: A large-scale hierarchical image database.
\newblock In \emph{CVPR}, 2009.

\bibitem[Guo et~al.(2019)Guo, Zhang, Mu, Heng, Liu, Wei, and
  Sun]{guo2019single}
Zichao Guo, Xiangyu Zhang, Haoyuan Mu, Wen Heng, Zechun Liu, Yichen Wei, and
  Jian Sun.
\newblock Single path one-shot neural architecture search with uniform
  sampling.
\newblock \emph{arXiv preprint arXiv:1904.00420}, 2019.

\bibitem[Han et~al.(2015)Han, Pool, Tran, and Dally]{han2015learning}
Song Han, Jeff Pool, John Tran, and William Dally.
\newblock Learning both weights and connections for efficient neural network.
\newblock In \emph{NeurIPS}, 2015.

\bibitem[Han et~al.(2016)Han, Mao, and Dally]{han2016deep}
Song Han, Huizi Mao, and William~J Dally.
\newblock Deep compression: Compressing deep neural networks with pruning,
  trained quantization and huffman coding.
\newblock In \emph{ICLR}, 2016.

\bibitem[Hao et~al.(2019)Hao, Zhang, Li, Huang, Xiong, Rupnow, Hwu, and
  Chen]{hao2019fpga}
Cong Hao, Xiaofan Zhang, Yuhong Li, Sitao Huang, Jinjun Xiong, Kyle Rupnow,
  Wen-mei Hwu, and Deming Chen.
\newblock Fpga/dnn co-design: An efficient design methodology for 1ot
  intelligence on the edge.
\newblock In \emph{2019 56th ACM/IEEE Design Automation Conference (DAC)}, pp.\
   1--6. IEEE, 2019.

\bibitem[He et~al.(2016)He, Zhang, Ren, and Sun]{he2016deep}
Kaiming He, Xiangyu Zhang, Shaoqing Ren, and Jian Sun.
\newblock Deep residual learning for image recognition.
\newblock In \emph{CVPR}, 2016.

\bibitem[He et~al.(2018)He, Lin, Liu, Wang, Li, and Han]{he2018amc}
Yihui He, Ji~Lin, Zhijian Liu, Hanrui Wang, Li-Jia Li, and Song Han.
\newblock Amc: Automl for model compression and acceleration on mobile devices.
\newblock In \emph{ECCV}, 2018.

\bibitem[Hinton et~al.(2015)Hinton, Vinyals, and Dean]{hinton2015distilling}
Geoffrey Hinton, Oriol Vinyals, and Jeff Dean.
\newblock Distilling the knowledge in a neural network.
\newblock \emph{arXiv preprint arXiv:1503.02531}, 2015.

\bibitem[Howard et~al.(2019)Howard, Sandler, Chu, Chen, Chen, Tan, Wang, Zhu,
  Pang, Vasudevan, et~al.]{howard2019searching}
Andrew Howard, Mark Sandler, Grace Chu, Liang-Chieh Chen, Bo~Chen, Mingxing
  Tan, Weijun Wang, Yukun Zhu, Ruoming Pang, Vijay Vasudevan, et~al.
\newblock Searching for mobilenetv3.
\newblock In \emph{ICCV 2019}, 2019.

\bibitem[Howard et~al.(2017)Howard, Zhu, Chen, Kalenichenko, Wang, Weyand,
  Andreetto, and Adam]{howard2017mobilenets}
Andrew~G Howard, Menglong Zhu, Bo~Chen, Dmitry Kalenichenko, Weijun Wang,
  Tobias Weyand, Marco Andreetto, and Hartwig Adam.
\newblock Mobilenets: Efficient convolutional neural networks for mobile vision
  applications.
\newblock \emph{arXiv preprint arXiv:1704.04861}, 2017.

\bibitem[Huang et~al.(2017)Huang, Liu, Van Der~Maaten, and
  Weinberger]{huang2017densely}
Gao Huang, Zhuang Liu, Laurens Van Der~Maaten, and Kilian~Q Weinberger.
\newblock Densely connected convolutional networks.
\newblock In \emph{CVPR}, 2017.

\bibitem[Huang et~al.(2018)Huang, Chen, Li, Wu, van~der Maaten, and
  Weinberger]{huang2017multi}
Gao Huang, Danlu Chen, Tianhong Li, Felix Wu, Laurens van~der Maaten, and
  Kilian~Q Weinberger.
\newblock Multi-scale dense networks for resource efficient image
  classification.
\newblock In \emph{ICLR}, 2018.

\bibitem[Iandola et~al.(2016)Iandola, Han, Moskewicz, Ashraf, Dally, and
  Keutzer]{iandola2016squeezenet}
Forrest~N Iandola, Song Han, Matthew~W Moskewicz, Khalid Ashraf, William~J
  Dally, and Kurt Keutzer.
\newblock Squeezenet: Alexnet-level accuracy with 50x fewer parameters and< 0.5
  mb model size.
\newblock \emph{arXiv preprint arXiv:1602.07360}, 2016.

\bibitem[Jiang et~al.(2019{\natexlab{a}})Jiang, Yang, Sha, Zhuge, Gu, Shi, and
  Hu]{jiang2019hardware}
Weiwen Jiang, Lei Yang, Edwin Sha, Qingfeng Zhuge, Shouzhen Gu, Yiyu Shi, and
  Jingtong Hu.
\newblock Hardware/software co-exploration of neural architectures.
\newblock \emph{arXiv preprint arXiv:1907.04650}, 2019{\natexlab{a}}.

\bibitem[Jiang et~al.(2019{\natexlab{b}})Jiang, Zhang, Sha, Yang, Zhuge, Shi,
  and Hu]{jiang2019accuracy}
Weiwen Jiang, Xinyi Zhang, Edwin H-M Sha, Lei Yang, Qingfeng Zhuge, Yiyu Shi,
  and Jingtong Hu.
\newblock Accuracy vs. efficiency: Achieving both through fpga-implementation
  aware neural architecture search.
\newblock In \emph{Proceedings of the 56th Annual Design Automation Conference
  2019}, pp.\  1--6, 2019{\natexlab{b}}.

\bibitem[Kuen et~al.(2018)Kuen, Kong, Lin, Wang, Yin, See, and
  Tan]{kuen2018stochastic}
Jason Kuen, Xiangfei Kong, Zhe Lin, Gang Wang, Jianxiong Yin, Simon See, and
  Yap-Peng Tan.
\newblock Stochastic downsampling for cost-adjustable inference and improved
  regularization in convolutional networks.
\newblock In \emph{CVPR}, 2018.

\bibitem[Lin et~al.(2017)Lin, Rao, Lu, and Zhou]{lin2017runtime}
Ji~Lin, Yongming Rao, Jiwen Lu, and Jie Zhou.
\newblock Runtime neural pruning.
\newblock In \emph{NeurIPS}, 2017.

\bibitem[Liu et~al.(2018)Liu, Zoph, Neumann, Shlens, Hua, Li, Fei-Fei, Yuille,
  Huang, and Murphy]{liu2018progressive}
Chenxi Liu, Barret Zoph, Maxim Neumann, Jonathon Shlens, Wei Hua, Li-Jia Li,
  Li~Fei-Fei, Alan Yuille, Jonathan Huang, and Kevin Murphy.
\newblock Progressive neural architecture search.
\newblock In \emph{ECCV}, 2018.

\bibitem[Liu et~al.(2019)Liu, Simonyan, and Yang]{liu2019darts}
Hanxiao Liu, Karen Simonyan, and Yiming Yang.
\newblock Darts: Differentiable architecture search.
\newblock In \emph{ICLR}, 2019.

\bibitem[Liu \& Deng(2018)Liu and Deng]{liu2018dynamic}
Lanlan Liu and Jia Deng.
\newblock Dynamic deep neural networks: Optimizing accuracy-efficiency
  trade-offs by selective execution.
\newblock In \emph{AAAI}, 2018.

\bibitem[Liu et~al.(2017)Liu, Li, Shen, Huang, Yan, and Zhang]{liu2017learning}
Zhuang Liu, Jianguo Li, Zhiqiang Shen, Gao Huang, Shoumeng Yan, and Changshui
  Zhang.
\newblock Learning efficient convolutional networks through network slimming.
\newblock In \emph{ICCV}, 2017.

\bibitem[Loshchilov \& Hutter(2016)Loshchilov and Hutter]{loshchilov2016sgdr}
Ilya Loshchilov and Frank Hutter.
\newblock Sgdr: Stochastic gradient descent with warm restarts.
\newblock \emph{arXiv preprint arXiv:1608.03983}, 2016.

\bibitem[Ma et~al.(2018)Ma, Zhang, Zheng, and Sun]{ma2018shufflenet}
Ningning Ma, Xiangyu Zhang, Hai-Tao Zheng, and Jian Sun.
\newblock Shufflenet v2: Practical guidelines for efficient cnn architecture
  design.
\newblock In \emph{ECCV}, 2018.

\bibitem[Real et~al.(2019)Real, Aggarwal, Huang, and Le]{real2018regularized}
Esteban Real, Alok Aggarwal, Yanping Huang, and Quoc~V Le.
\newblock Regularized evolution for image classifier architecture search.
\newblock In \emph{AAAI}, 2019.

\bibitem[Sandler et~al.(2018)Sandler, Howard, Zhu, Zhmoginov, and
  Chen]{sandler2018mobilenetv2}
Mark Sandler, Andrew Howard, Menglong Zhu, Andrey Zhmoginov, and Liang-Chieh
  Chen.
\newblock Mobilenetv2: Inverted residuals and linear bottlenecks.
\newblock In \emph{CVPR}, 2018.

\bibitem[Strubell et~al.(2019)Strubell, Ganesh, and
  McCallum]{strubell2019energy}
Emma Strubell, Ananya Ganesh, and Andrew McCallum.
\newblock Energy and policy considerations for deep learning in nlp.
\newblock In \emph{ACL}, 2019.

\bibitem[Tan et~al.(2019)Tan, Chen, Pang, Vasudevan, Sandler, Howard, and
  Le]{tan2018mnasnet}
Mingxing Tan, Bo~Chen, Ruoming Pang, Vijay Vasudevan, Mark Sandler, Andrew
  Howard, and Quoc~V Le.
\newblock Mnasnet: Platform-aware neural architecture search for mobile.
\newblock In \emph{Proceedings of the IEEE Conference on Computer Vision and
  Pattern Recognition}, pp.\  2820--2828, 2019.

\bibitem[Wang et~al.(2018)Wang, Yu, Dou, Darrell, and
  Gonzalez]{wang2018skipnet}
Xin Wang, Fisher Yu, Zi-Yi Dou, Trevor Darrell, and Joseph~E Gonzalez.
\newblock Skipnet: Learning dynamic routing in convolutional networks.
\newblock In \emph{ECCV}, 2018.

\bibitem[Williams et~al.(2009)Williams, Waterman, and
  Patterson]{williams2009roofline}
Samuel Williams, Andrew Waterman, and David Patterson.
\newblock Roofline: An insightful visual performance model for floating-point
  programs and multicore architectures.
\newblock Technical report, Lawrence Berkeley National Lab.(LBNL), Berkeley, CA
  (United States), 2009.

\bibitem[Wu et~al.(2019)Wu, Dai, Zhang, Wang, Sun, Wu, Tian, Vajda, Jia, and
  Keutzer]{wu2018fbnet}
Bichen Wu, Xiaoliang Dai, Peizhao Zhang, Yanghan Wang, Fei Sun, Yiming Wu,
  Yuandong Tian, Peter Vajda, Yangqing Jia, and Kurt Keutzer.
\newblock Fbnet: Hardware-aware efficient convnet design via differentiable
  neural architecture search.
\newblock In \emph{CVPR}, 2019.

\bibitem[Wu et~al.(2018)Wu, Nagarajan, Kumar, Rennie, Davis, Grauman, and
  Feris]{wu2018blockdrop}
Zuxuan Wu, Tushar Nagarajan, Abhishek Kumar, Steven Rennie, Larry~S Davis,
  Kristen Grauman, and Rogerio Feris.
\newblock Blockdrop: Dynamic inference paths in residual networks.
\newblock In \emph{CVPR}, 2018.

\bibitem[Yu \& Huang(2019{\natexlab{a}})Yu and Huang]{yu2019autoslim}
Jiahui Yu and Thomas Huang.
\newblock Autoslim: Towards one-shot architecture search for channel numbers.
\newblock \emph{arXiv preprint arXiv:1903.11728}, 2019{\natexlab{a}}.

\bibitem[Yu \& Huang(2019{\natexlab{b}})Yu and Huang]{yu2019universally}
Jiahui Yu and Thomas Huang.
\newblock Universally slimmable networks and improved training techniques.
\newblock In \emph{ICCV}, 2019{\natexlab{b}}.

\bibitem[Yu et~al.(2019)Yu, Yang, Xu, Yang, and Huang]{yu2018slimmable}
Jiahui Yu, Linjie Yang, Ning Xu, Jianchao Yang, and Thomas Huang.
\newblock Slimmable neural networks.
\newblock In \emph{ICLR}, 2019.

\bibitem[Zhang et~al.(2018)Zhang, Zhou, Lin, and Sun]{zhang2018shufflenet}
Xiangyu Zhang, Xinyu Zhou, Mengxiao Lin, and Jian Sun.
\newblock Shufflenet: An extremely efficient convolutional neural network for
  mobile devices.
\newblock In \emph{CVPR}, 2018.

\bibitem[Zhu et~al.(2017)Zhu, Han, Mao, and Dally]{zhu2016trained}
Chenzhuo Zhu, Song Han, Huizi Mao, and William~J Dally.
\newblock Trained ternary quantization.
\newblock In \emph{ICLR}, 2017.

\bibitem[Zoph \& Le(2017)Zoph and Le]{zoph2017neural}
Barret Zoph and Quoc~V Le.
\newblock Neural architecture search with reinforcement learning.
\newblock In \emph{ICLR}, 2017.

\bibitem[Zoph et~al.(2018)Zoph, Vasudevan, Shlens, and Le]{zoph2018learning}
Barret Zoph, Vijay Vasudevan, Jonathon Shlens, and Quoc~V Le.
\newblock Learning transferable architectures for scalable image recognition.
\newblock In \emph{CVPR}, 2018.

\end{thebibliography}
\bibliographystyle{iclr2020_conference}


\appendix
\section{Details of the Accuracy Predictor}\label{appendix:acc_predictor}
\begin{figure}[ht]
    \centering
    \includegraphics[width=0.45\textwidth]{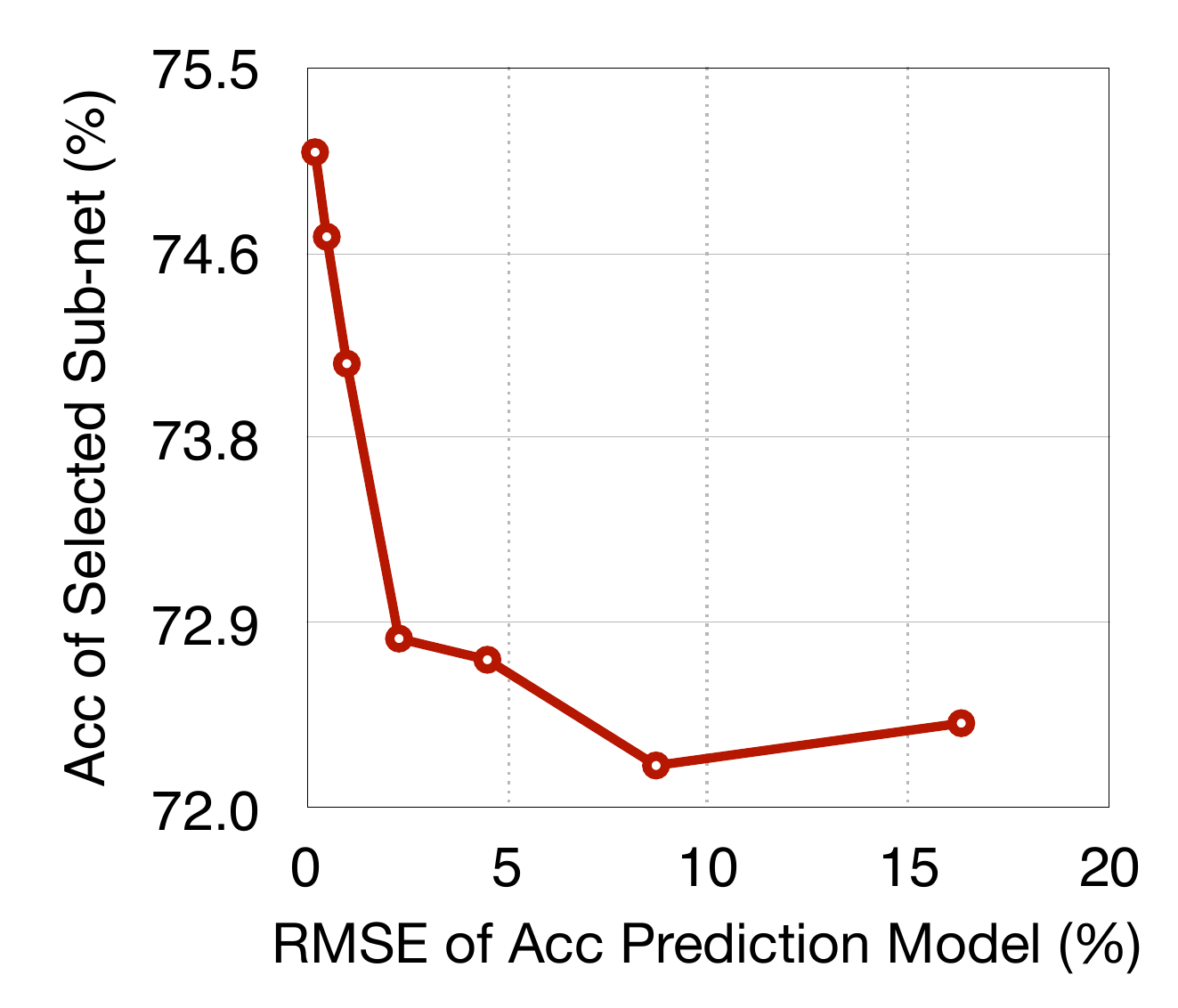}
    \caption{Performances of selected sub-networks using different accuracy prediction model.}
    \label{fig:acc_predictor_analysis}
\end{figure}

We use a three-layer feedforward neural network that has 400 hidden units in each layer as the accuracy predictor. Given a model, we encode each layer in the neural network into a one-hot vector based on its kernel size and expand ratio, and we assign zero vectors to layers that are skipped. Besides, we have an additional one-hot vector that represents the input image size. We concatenate these vectors into a large vector that represents the whole neural network architecture and input image size, which is then fed to the three-layer feedforward neural network to get the predicted accuracy. In our experiments, this simple accuracy prediction model can provide very accurate predictions. At convergence, the root-mean-square error (RMSE) between predicted accuracy and estimated accuracy on the test set is only 0.21\%. Figure~\ref{fig:acc_predictor_analysis} shows the relationship between the RMSE of the accuracy prediction model and the final results (i.e., the accuracy of selected sub-networks). We can find that lower RMSE typically leads to better final results. 

\section{Implementation Details of Progressive Shrinking}\label{appendix:ps_details}
After training the full network, we first have one stage of fine-tuning to incorporate elastic kernel size. In this stage (i.e., $K \in [7, 5, 3]$), we sample one sub-network in each update step. The network is fine-tuned for 125 epochs with an initial learning rate of 0.96. All other training settings are the same as training the full network. 

Next, we have two stages of fine-tuning to incorporate elastic depth. We sample two sub-networks and aggregate their gradients in each update step. The first stage (i.e., $D \in [4, 3]$) takes 25 epochs with an initial learning rate of 0.08 while the second stage (i.e., $D \in [4, 3, 2]$) takes 125 epochs with an initial learning rate of 0.24.

Finally, we have two stages of fine-tuning to incorporate elastic width. We sample four sub-networks and aggregate their gradients in each update step. The first stage (i.e., $W \in [6, 4]$) takes 25 epochs with an initial learning rate of 0.08 while the second stage (i.e., $W \in [6, 4, 3]$) takes 125 epochs with an initial learning rate of 0.24.

\end{document}